\documentclass{article}

\usepackage[preprint]{neurips_2026}

\usepackage[utf8]{inputenc} 
\usepackage[T1]{fontenc}    
\usepackage{hyperref}       
\usepackage{url}            
\usepackage{booktabs}       
\usepackage{amsfonts}       
\usepackage{nicefrac}       
\usepackage{microtype}      
\usepackage{xcolor}         
\usepackage{algorithm}
\usepackage{algorithmic}
\usepackage{amsmath}
\usepackage{tabularx}
\usepackage{array}
\usepackage{booktabs}
\usepackage{multirow}
\usepackage{graphicx}
\usepackage{adjustbox}
\usepackage{amsmath}
\usepackage{amssymb}
\usepackage[most]{tcolorbox}

\title{
\makebox[0pt][r]{%
    \raisebox{-0.5\height}[0pt][0pt]{%
        \includegraphics[height=2.2cm]{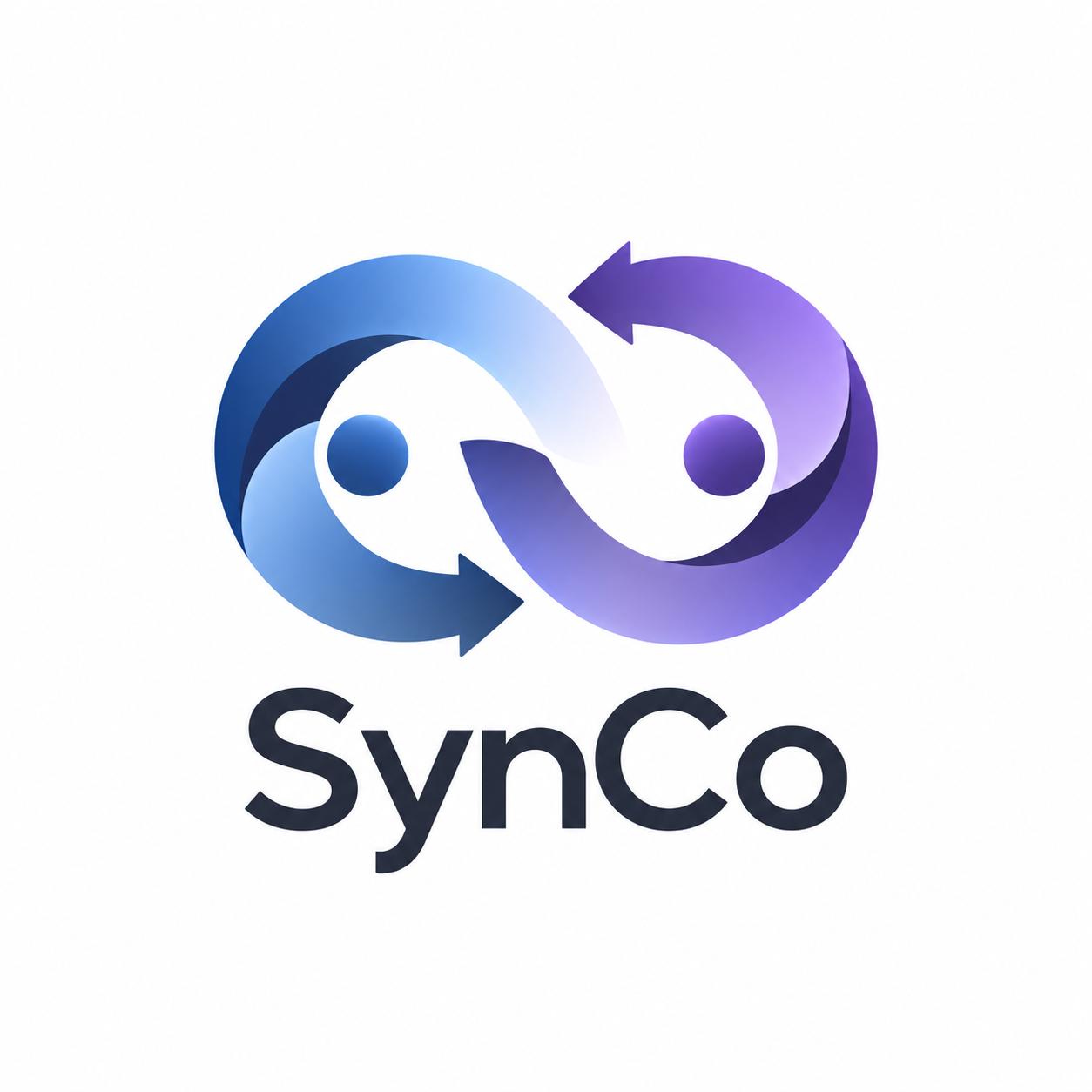}
    }\hspace{0.6cm}%
}%
SynCo: Data Synthesis Co-Training for Self-Evolving LLMs via Multi-Agent Reinforcement Learning
}

\author{%
  \textbf{Wei Yang$^{1}$, Shawn Li$^{1}$, Yuehan Qin$^{1}$, Yawei Wang$^{2}$, Mingxi Wang$^{1}$, Shixuan Li$^{1}$} \\
  \textbf{Tiankai Yang$^{1}$, Jiate Li$^{1}$, Jesse Thomason$^{3}$, Xuezhe Ma$^{1}$, Yue Zhao$^{1}$} \\
  $^{1}$University of Southern California \\
  $^{2}$ The George Washington University \\
  $^{3}$Georgia Institute of Technology \\
}

\begin{document}

\maketitle

\begin{abstract}
Self-evolving LLM agents promise to improve autonomously through continual interaction and learning, reducing their dependence on manually curated supervision. Realizing this promise requires not only updating the agent, but also evolving its training experience as its capabilities change. However, most existing pipelines rely on static datasets or separately updated synthesis models, causing previously useful tasks to become trivial while overly difficult tasks remain uninformative. This growing mismatch between agent capability and training experience limits sustained self-improvement.
To address this problem, we propose \textsc{SynCo}, an agentic data synthesis co-training framework for self-evolving LLMs based on multi-agent reinforcement learning.
\textsc{SynCo} jointly optimizes two independently parameterized agents: a Synthesizer that constructs training tasks from the Reasoner's evolving capability state, and a Reasoner that learns from the resulting experience. Each synthesized task induces multiple Reasoner rollouts whose outcomes provide complementary rewards to both agents. Correctness feedback improves the Reasoner, while task quality, answer reliability, and outcome-grounded teachability guide the Synthesizer. Their updates are fed back into subsequent synthesis rounds, allowing the task-solving policy and its training distribution to evolve together. Extensive experiments across eight mathematical reasoning benchmarks demonstrate that \textsc{SynCo} substantially outperforms a broad range of existing synthetic-data methods and controlled baselines, achieving the strongest overall performance while deriving most of its gains from previously unsolved problems. \\
GitHub: \url{https://github.com/weiyang930/SynCo.git}\\
Hugging Face: \url{https://huggingface.co/datasets/weiyang930/SynCo}
\end{abstract}

\section{Introduction}
\label{sec:introduction}

Large language model agents are increasingly expected to improve through interaction and accumulated experience, rather than remain bounded by a fixed collection of human-curated demonstrations~\citep{tang2025synthesizing,gao2026self}.
Yet most agent training pipelines still rely on data that is static, externally constructed, or only periodically refreshed. This creates a fundamental mismatch: agent capabilities change throughout training, while the distribution of learning experience does not~\citep{pahuja2025explorer,liang2026sws}. Tasks that were once informative may become trivial, whereas tasks far beyond the agent's competence continue to provide little useful learning signal. The resulting misalignment between the learner and its data limits both training efficiency and sustained improvement. A genuinely self-evolving agent therefore requires not merely more synthetic data, but a source of training experience that continuously adapts to its evolving capabilities~\citep{cheng2025evolving,zhang2026evofsm,yue2026dr}.

Recent work has begun to address this limitation through adaptive data synthesis, weakness-driven task generation, and teacher--solver or challenger--solver co-evolution~\citep{zhai2025agentevolver,yang2026coevolve,yue2026dr}. These approaches adapt the training data to the current learner by regenerating, selecting, or reweighting tasks according to model failures, uncertainty, or estimated difficulty. However, an adaptive data distribution does not necessarily imply joint optimization of data synthesis and task learning.
Many existing frameworks still alternate between generating tasks from a previous learner checkpoint and subsequently training the learner on the resulting data~\citep{huang2025r,wang2025socratic}.
Although this enables iterative adaptation, the synthesis policy and the task policy are not optimized together from the same online interaction. Data generation therefore remains an auxiliary process around agent training rather than a learnable component of the training system itself. True mutual evolution requires jointly optimizing the experience synthesizer and the learner.

Such joint training creates a coupled and non-stationary learning problem. The Synthesizer shapes the Reasoner's optimization landscape through the tasks it generates, while each Reasoner update redefines which tasks remain useful. A task that is informative at one stage may later become trivial, whereas a nominally difficult task may still provide little supervision if the Reasoner consistently succeeds or fails. The value of synthetic data therefore cannot be judged solely by static properties such as validity, novelty, or nominal difficulty, but must be grounded in the current Reasoner's actual interaction with the data~\citep{yang2026coevolve,kwan2026scope}. This leads to the central question of our work: \emph{Can agentic data synthesis and task learning be formulated as a unified multi-agent reinforcement learning process in which the data-synthesis policy and the task-solving policy are jointly optimized and continuously evolve together?} Answering this question requires more than connecting a generator to a learner. It requires a closed loop that integrates learner-state observation, adaptive task generation, shared interaction feedback, and online updates of both policies.

To address this challenge, we propose \textsc{SynCo}, an agentic data synthesis co-training framework for self-evolving LLMs based on multi-agent reinforcement learning. \textsc{SynCo} jointly optimizes two independently parameterized agents: a \emph{Synthesizer} that constructs training tasks and a \emph{Reasoner} that learns from them.
At each iteration, the Synthesizer observes an evolving context built from the recent synthetic data pool, the Reasoner's empirical capability profile, and feedback from rejected tasks. It then generates a structured and verifiable task tailored to the Reasoner's current learning needs. The Reasoner performs multiple solution rollouts, whose outcomes provide complementary learning signals to both agents. Correctness rewards improve the Reasoner, while task quality, answer reliability, and outcome-grounded teachability guide the Synthesizer. Both policies are updated online through group-relative policy optimization within the same multi-agent training process. Their updates define the next synthesis round: the improved Reasoner induces a new capability state, and the improved Synthesizer converts that state into a new training distribution.
Through this closed loop, \textsc{SynCo} enables the task-solving policy and its training data to evolve together.

Our main contributions are summarized as follows:
\begin{itemize}
    \item We formulate agentic data synthesis and task learning as a coupled multi-agent reinforcement learning problem, and propose \textsc{SynCo} to jointly optimize a Synthesizer and a Reasoner within a unified online training process.

    \item We develop a closed-loop self-evolution framework in which the Synthesizer continually adapts the training distribution to the evolving Reasoner, while outcome-grounded interaction feedback from the Reasoner directly supervises the synthesis policy.

    \item Extensive experiments across diverse mathematical reasoning benchmarks show that \textsc{SynCo} consistently outperforms a frozen-Synthesizer baseline, with most gains arising from previously unsolved problems.
\end{itemize}

\section{Related Work}
\label{sec:related_work}

Synthetic data has become central to LLM post-training, ranging from instruction bootstrapping methods such as Self-Instruct~\citep{wang2023self} and WizardLM~\citep{xu2024wizardlm} to reasoning-oriented pipelines such as STaR~\citep{zelikman2022star}, MetaMath~\citep{yu2023metamath}, and DeepSeekMath~\citep{shao2024deepseekmath}.
Recent work further improves realism, diversity, verifiability, and learner relevance: MATRIX~\citep{tang2025synthesizing} generates scenarios through multi-agent simulation, FAC Synthesis~\citep{li2026less} targets missing representation-space features, Explorer~\citep{pahuja2025explorer} synthesizes trajectories through environment exploration, and SwS~\citep{liang2026sws} constructs problems around persistent learner weaknesses.
Related pipelines extend synthetic supervision to mathematical reasoning, tool use, web interaction, and executable environments~\citep{gou2024tora,ramrakhya2025scaling,wang2026trajectory2task,chen2026eigendata}.
Most of these approaches nevertheless construct an offline corpus or adapt generation separately from downstream policy optimization.

Self-evolving frameworks more directly couple experience generation with learner improvement.
R-Zero~\citep{huang2025r}, Socratic-Zero~\citep{wang2025socratic}, and Agent0~\citep{xia2025agent0} establish co-evolving proposer, teacher, or curriculum roles for autonomous reasoning improvement.
Dr. Zero~\citep{yue2026dr} and SCOPE~\citep{kwan2026scope} extend this paradigm to multi-turn search and open-ended document-grounded tasks, while CoEvolve~\citep{yang2026coevolve} adapts tool-use task distributions using rollout-derived forgetting and uncertainty signals.
Agent-World~\citep{dong2026agent}, SEAD~\citep{dai2026sead}, AgentEvolver~\citep{zhai2025agentevolver}, EvolveR~\citep{wu2025evolver}, and EvoFSM~\citep{zhang2026evofsm} further explore the evolution of environments, curricula, memories, experiences, and agent workflows.
A comprehensive discussion of synthetic-data generation and self-evolving agents is provided in Appendix.A.

\section{Method}
\label{sec:method}

\begin{figure*}[t]
    \centering
    \includegraphics[width=0.95\textwidth]{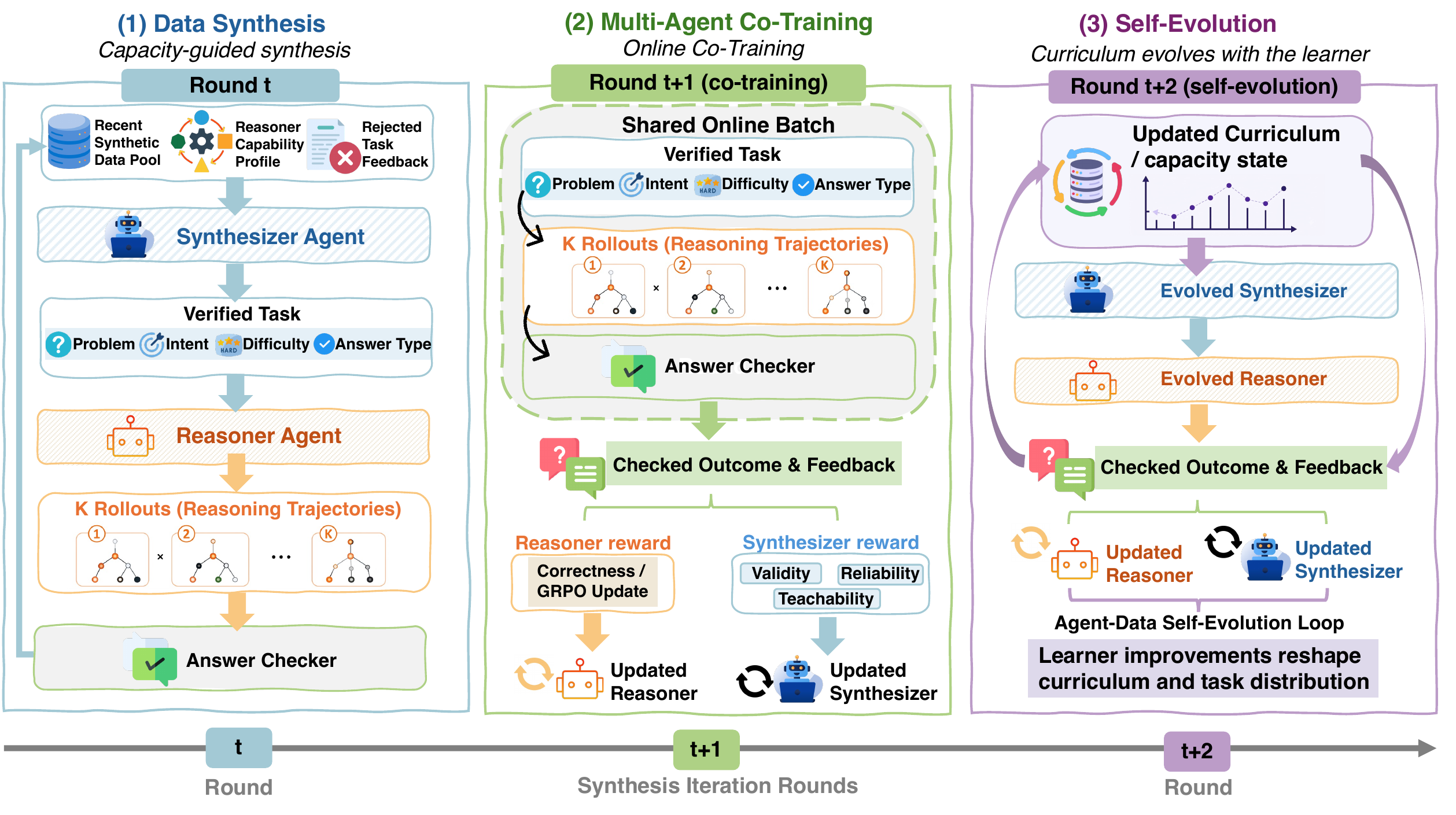}
    \caption{
    Overview of \textsc{SynCo}, a closed-loop framework for self-evolving LLMs.
    (1) A Synthesizer first generates capability-guided and verifiable tasks from the current learning context.
    (2) The resulting interaction batch is then used to jointly update the Synthesizer and the Reasoner through online multi-agent co-training.
    (3) The updated agents in turn reshape the future curriculum and task distribution, forming an agent--data self-evolution loop.
    }
    \label{fig:synco_framework}
\end{figure*}

In this section, we present \textsc{SynCo}, a multi-agent reinforcement learning framework that jointly optimizes agentic data synthesis and task reasoning for continual self-evolution.
The overall framework is illustrated in Figure~\ref{fig:synco_framework}.

\subsection{Problem Formulation: Agentic Data Synthesis Co-Training}
\label{sec:problem_formulation}

We consider two independently parameterized agents: a \emph{Synthesizer}
$\pi_S(\cdot;\theta_S)$ that generates training tasks and a \emph{Reasoner}
$\pi_R(\cdot;\theta_R)$ that learns to solve them.
The Synthesizer determines the Reasoner's training experience, while the Reasoner's realized behavior provides feedback for improving synthesis.

At iteration $t$, a dynamic context $c_t$ summarizes the Reasoner's recent capabilities, the coverage of previously generated tasks, and feedback from rejected samples.
Conditioned on $c_t$, the Synthesizer generates a structured and verifiable task:
\begin{equation}
    x_t \sim \pi_S(\cdot\mid c_t;\theta_S^t),
    \qquad
    \mathcal{D}_t=\pi_S(\cdot\mid c_t;\theta_S^t),
    \label{eq:dynamic_synthesis}
\end{equation}
where $\mathcal{D}_t$ is the online training distribution induced by the current learner state and synthesis policy.

For each admitted task $x_t$ with verified answer $a_t^\star$, the Reasoner samples $K$ independent rollouts
$y_t^{(k)}\sim\pi_R(\cdot\mid x_t;\theta_R^t)$, whose outcomes are determined by
$z_t^{(k)}=\operatorname{Check}(y_t^{(k)},a_t^\star)$.
The same interaction yields rollout-level correctness rewards $r_{R,t}^{(k)}$ for the Reasoner and a task-level reward $r_{S,t}$ that evaluates both synthesis quality and realized utility for the current Reasoner.

The agents optimize distinct but coupled objectives:
\begin{equation}
    J_R(\theta_R)
    =
    \mathbb{E}\!\left[
        \frac{1}{K}\sum_{k=1}^{K}r_{R,t}^{(k)}
    \right],
    \qquad
    J_S(\theta_S)
    =
    \mathbb{E}\!\left[r_{S,t}\right].
    \label{eq:coupled_objectives}
\end{equation}
The objectives are coupled because $\pi_S$ determines the distribution underlying $J_R$, whereas $\pi_R$ produces the interaction feedback underlying $J_S$.
Both policies are updated from the same online batch using agent-specific group-relative policy optimization, forming a unified co-training process rather than sequential synthesis and learning.

\subsection{Capability-Conditioned Agentic Data Synthesis}
\label{sec:agentic_synthesis}

A self-evolving Reasoner requires training experience that remains aligned with its changing capabilities.
Rather than sampling from a fixed generator, the Synthesizer observes recent agent--data interactions, identifies which capabilities should be exercised next, and realizes this decision as a structured and verifiable task.
Data generation therefore becomes an adaptive policy conditioned on the evolving learner.

\paragraph{Capability-conditioned synthesis context.}
Let $\mathcal{M}_t$ denote the interaction memory before iteration $t$, containing recent tasks, validation outcomes, Reasoner rollouts and success rates, synthesis rewards, and rejection records.
We summarize it as
\begin{equation}
    c_t
    =
    G(\mathcal{M}_t)
    =
    \left(p_t,d_t,e_t\right),
    \label{eq:synthesis_context}
\end{equation}
where $p_t$ characterizes the current synthetic pool and its coverage, $d_t$ summarizes the Reasoner's empirical performance and remaining weaknesses, and $e_t$ records synthesis failures, redundancy, and under-explored task regions.
The context is reconstructed throughout training, allowing synthesis to adapt as both the Reasoner and the accumulated task distribution evolve.

\paragraph{Structured pedagogical task generation.}
Conditioned on $c_t$, the Synthesizer generates a structured task card
\begin{equation}
    x_t
    =
    \left(
        q_t,\mathbf{m}_t,s_t^\star,a_t^\star
    \right)
    \sim
    \pi_S(\cdot\mid c_t;\theta_S^t),
    \label{eq:structured_task}
\end{equation}
where $q_t$ is the problem presented to the Reasoner, $\mathbf{m}_t$ specifies its pedagogical intent and estimated difficulty, and $(s_t^\star,a_t^\star)$ provides a reference solution and final answer for verification.
The metadata encourages an explicit synthesis decision by specifying the targeted reasoning behavior, anticipated weakness, and novelty or verification requirements.
Accordingly, the Synthesizer is instructed to produce tasks that are original, self-contained, non-trivial, and aligned with weaknesses or under-explored capabilities identified in $c_t$.

The reference solution, answer, and metadata are used only for validation and reward computation.
The Reasoner receives $q_t$ alone, preventing answer leakage and ensuring that its rollout outcomes reflect its own policy.

\paragraph{Verified admission.}
Because open-ended generation may produce malformed, ambiguous, inconsistent, or repetitive tasks, every task is evaluated by a verifier
$V(x_t)=(v_t,\boldsymbol{\rho}_t)$.
Here, $\boldsymbol{\rho}_t$ records schema validity, problem completeness, answer consistency, output format, and duplication, while $v_t$ determines whether the task enters Reasoner training:
\begin{equation}
    v_t
    =
    \mathbb{I}\!\left[x_t\in\mathcal{X}_{\mathrm{valid}}\right],
    \
    \widetilde{\mathcal{D}}_t
    =
    \left\{
        x_t\sim\pi_S(\cdot\mid c_t;\theta_S^t)
        \mid v_t=1
    \right\}.
    \label{eq:task_admission}
\end{equation}
When enabled, an additional consensus check independently solves the generated problem and rejects tasks whose inferred answers do not support the Synthesizer-provided reference.
Invalid tasks are excluded from Reasoner training and return negative feedback to the Synthesizer, whereas admitted tasks form the adaptive distribution used for subsequent multi-agent optimization.

\subsection{Interaction-Grounded Multi-Agent Co-Training}
\label{sec:marl_cotraining}

The verified tasks provide a shared interaction space for both agents.
For each admitted task, the Reasoner's rollouts simultaneously provide direct supervision for task solving and outcome-grounded feedback for data synthesis.
This shared interaction couples the two policies within the same online training process.

\paragraph{Reasoner interaction and outcome verification.}
Given a valid task $x_t$, the Reasoner receives only its problem statement $q_t$ and samples $K$ independent trajectories:
\begin{equation}
\begin{aligned}
    y_t^{(k)}
    &\sim
    \pi_R(\cdot\mid q_t;\theta_R^t),\\
    z_t^{(k)}
    &=
    \operatorname{Check}
    \left(
        \hat{a}_t^{(k)},a_t^\star
    \right)
    \in\{0,1\},\\
    \mu_t
    &=
    \frac{1}{K}
    \sum_{k=1}^{K}z_t^{(k)},
    \qquad k=1,\ldots,K,
\end{aligned}
\label{eq:group_outcomes}
\end{equation}
where $\hat{a}_t^{(k)}$ is the extracted answer and $\mu_t$ is the empirical success rate of the current Reasoner on task $x_t$.
The checker supports normalized exact, numerical, symbolic, and set- or tuple-level equivalence.
The resulting rollout group distinguishes uniformly solved or failed tasks from those producing mixed outcomes, making $\mu_t$ an outcome-grounded measure of task utility for the current Reasoner.

\paragraph{Agent-specific interaction rewards.}
The Reasoner is optimized with rollout-level correctness rewards, encouraging trajectories that produce verified answers.
The Synthesizer instead receives a task-level gated reward:
\begin{equation}
    r_{S,t}
    =
    G_q\!\left(r_t^{\mathrm{quality}}\right)
    G_r\!\left(r_t^{\mathrm{reliability}}\right)
    T(\mu_t),
    \label{eq:synthesizer_reward}
\end{equation}
where $G_q$ and $G_r$ softly gate task quality and answer reliability, while $T(\mu_t)$ measures realized teachability from the Reasoner's rollout outcomes.
The teachability signal is maximized for tasks producing mixed success and decreases toward uniformly solved or failed tasks, which provide little group-relative learning signal.
Thus, quality and reliability constrain the feasible synthesis space, while the current Reasoner's behavior determines which valid tasks are most useful for learning.
Exact reward definitions and implementation constants are provided in Appendix~\ref{app:reward_details}.

\paragraph{Joint multi-agent update.}
At each iteration, the system collects a shared online interaction batch
\begin{equation}
    \mathcal{B}_t
    =
    \left\{
        \left(
            c_i,\,
            x_i,\,
            \left\{
                y_i^{(k)},z_i^{(k)}
            \right\}_{k=1}^{K}
        \right)
    \right\}_{i=1}^{B}.
    \label{eq:shared_interaction_batch}
\end{equation}
The same interaction batch induces separate group-relative objectives for the two agents: rollout-level correctness optimizes the Reasoner, while gated task utility optimizes the Synthesizer.
Although their parameters and rewards remain separate, both policies are updated within the same training iteration:
\begin{equation}
    \theta_R^{t+1}
    =
    \theta_R^t
    -
    \eta_R
    \nabla_{\theta_R}
    \mathcal{L}_R^{t},
    \qquad
    \theta_S^{t+1}
    =
    \theta_S^t
    -
    \eta_S
    \nabla_{\theta_S}
    \mathcal{L}_S^{t}.
    \label{eq:joint_policy_update}
\end{equation}
The Synthesizer determines the task distribution from which the Reasoner learns, while the Reasoner's realized outcomes determine the learning signal for the Synthesizer.
Updating either policy therefore changes the other's subsequent learning problem, forming a coupled online co-training process.

\subsection{Closed-Loop Agent--Data Self-Evolution}
\label{sec:self_evolution}

The co-training procedure in Section~\ref{sec:marl_cotraining} updates both the Reasoner and the policy that generates its training experience.
To sustain self-evolution, each interaction batch is fed back into a persistent memory that reconstructs the learner state and conditions the next synthesis round.

\paragraph{Feedback-driven context renewal.}
After processing $\mathcal{B}_t$, the system records the generated tasks, validation results, verified rollouts, empirical success rates, and agent-specific rewards.
Rejected tasks and their failure reasons are also retained to discourage repeated malformed, unreliable, or redundant generations.
The memory and synthesis context are updated as
\begin{equation}
\begin{aligned}
    \mathcal{M}_{t+1}
    &=
    \operatorname{Update}
    \left(
        \mathcal{M}_t,\mathcal{B}_t
    \right),\\
    c_{t+1}
    &=
    G\!\left(
        \operatorname{Recent}
        \left(
            \mathcal{M}_{t+1}
        \right)
    \right).
\end{aligned}
\label{eq:memory_context_update}
\end{equation}
The recent interaction window emphasizes the Reasoner's current behavior while retaining sufficient history to summarize task coverage, recurring weaknesses, and synthesis failures.
Thus, the synthesis condition evolves from realized learning behavior: previously challenging tasks may become trivial, new weaknesses may emerge, and over-produced reasoning patterns may be suppressed.

\paragraph{Coupled evolution of agents and data.}
As the Reasoner changes, the utility of a task becomes non-stationary:
$\mu(x;\pi_R^{t+1})\neq\mu(x;\pi_R^t)$.
A task that once produced mixed outcomes may later become uniformly solvable, while another may enter the learnable region as prerequisite capabilities improve.
Meanwhile, the updated Synthesizer changes how the renewed learner state is translated into future tasks.
The next verified distribution is therefore
\begin{equation}
    \widetilde{\mathcal{D}}_{t+1}
    =
    \left\{
        x
        \sim
        \pi_S
        \left(
            \cdot
            \mid
            c_{t+1};
            \theta_S^{t+1}
        \right)
        \;\middle|\;
        V(x).v=1
    \right\}.
    \label{eq:next_verified_distribution}
\end{equation}
Compared with $\widetilde{\mathcal{D}}_t$, this distribution changes through both a renewed context and an updated synthesis policy.
\textsc{SynCo} therefore does not merely query a fixed generator with a revised prompt; it jointly adapts the state conditioning generation and the policy that performs it.

\begin{table*}[t]
\centering
\fontsize{8.5}{9.5}\selectfont
\setlength{\tabcolsep}{3.3pt}
\renewcommand{\arraystretch}{1.06}
\caption{Reasoning accuracy (\%) across eight mathematical reasoning benchmarks. Vanilla denotes the initial Qwen3-8B model. Frozen Synthesizer retains online task generation but disables synthesis-policy updates. \emph{Avg.} is the unweighted average across all benchmarks. Bold and underlined values indicate the best and second-best results, respectively.}
\label{tab:rq1_main_results}
\begin{tabular}{lccccccccc}
\toprule
\textbf{Method} & \multicolumn{5}{c}{\textbf{Primary Benchmarks}} & \multicolumn{3}{c}{\textbf{Advanced Benchmarks}} & \textbf{Avg.} \\
\cmidrule(lr){2-6}\cmidrule(lr){7-9}
& \textbf{GSM8K} & \textbf{SVAMP} & \textbf{ASDiv} & \textbf{GSM-Hard} & \textbf{MATH} & \textbf{AIME24} & \textbf{AIME25} & \textbf{Minerva} & \\
\midrule
Vanilla & 92.42 & 84.00 & 95.98 & 51.78 & 81.20 & 73.33 & \underline{66.67} & 33.46 & 72.36 \\
\midrule
EnvScaler & 93.10 & 82.67 & 96.47 & 51.18 & 82.40 & 73.33 & 60.00 & \textbf{35.66} & 71.85 \\
AgenticQwen & \underline{95.00} & 89.33 & 97.29 & 55.04 & \textbf{88.00} & 60.00 & 60.00 & 30.88 & 71.94 \\
AgentSkiller & 83.17 & 71.00 & 92.45 & 39.88 & 73.60 & 66.67 & 56.67 & 31.62 & 64.38 \\
Klear-AgentForge & 91.21 & 89.67 & 97.70 & 55.42 & 75.20 & 26.67 & 13.33 & 27.21 & 59.55 \\
\midrule
MetaMathQA & 94.84 & 87.00 & 96.72 & 55.42 & 84.00 & 70.00 & \underline{66.67} & 34.56 & 73.65 \\
ScaleQuest-Math & 94.54 & 89.67 & 97.62 & 57.39 & 84.00 & \underline{76.67} & \underline{66.67} & 33.09 & 74.96 \\
OpenMathInstruct-2 & 94.54 & 89.00 & 97.45 & 57.62 & 86.40 & 70.00 & 56.67 & 33.46 & 73.14 \\
\midrule
Reject-Sample SFT & 92.49 & 83.67 & 96.14 & 51.48 & 82.40 & 73.33 & \underline{66.67} & 31.62 & 72.23 \\
Static-Pool GRPO & 92.95 & 85.00 & 96.06 & 52.46 & 82.20 & 70.00 & 63.33 & 31.99 & 71.75 \\
Off-Augment GRPO & 92.42 & 81.33 & 96.06 & 51.71 & 80.40 & \textbf{83.33} & \underline{66.67} & 33.09 & 73.13 \\
R-Zero & 92.80 & 84.67 & 96.31 & 51.55 & 83.20 & 80.00 & 66.67 & 33.46 & 73.58\\
\midrule
\textsc{SynCo} (Frozen Syn) & 94.77 & \underline{91.33} & \underline{97.78} & \underline{61.26} & 85.60 & \underline{76.67} & \underline{66.67} & 32.72 & \underline{75.85} \\
\textsc{SynCo} & \textbf{95.75} & \textbf{93.00} & \textbf{97.95} & \textbf{63.31} & \underline{87.80} & \textbf{83.33} & \textbf{73.33} & \underline{35.29} & \textbf{78.72} \\
\bottomrule
\end{tabular}
\end{table*}

\section{Experiments}
\label{sec:experiments}

\textbf{Benchmarks.}
We evaluate mathematical reasoning on eight benchmarks: GSM8K, SVAMP, ASDiv, GSM-Hard, MATH-500, AIME24, AIME25, and Minerva, reporting answer accuracy under a unified verification protocol.
\textbf{Baselines.}
We compare \textsc{SynCo} against existing synthetic-data methods, including AgenticQwen, EnvScaler, AgentSkiller, Klear-AgentForge, MetaMathQA, ScaleQuest-Math, OpenMathInstruct-2, and R-Zero, together with three controlled alternatives: Static-Pool GRPO, Offline-Augment GRPO, and Reject-Sample SFT.
We additionally include \textsc{SynCo} (Frozen Synth), which retains learner-conditioned online synthesis but disables Synthesizer updates.
\textbf{Implementation Details.}
We use Qwen3-8B as the default backbone and additionally evaluate Qwen3-4B for cross-scale validation.
Training is implemented with the UnityMAS-O multi-agent RL framework using GRPO.
Complete benchmark descriptions, training configurations, synthesis prompts, hyperparameters, and evaluation details are provided in Appendix~\ref{app:experiments}.

\subsection{RQ1: Does Co-Training Data Synthesis Improve Reasoning Performance?}
\label{sec:rq1}

Table~\ref{tab:rq1_main_results} presents the overall reasoning performance across eight mathematical benchmarks.
Across the broader comparison, existing synthetic-data methods achieve substantially lower average performance than \textsc{SynCo}; among them, ScaleQuest-Math is the strongest with 74.96\%, while the controlled Static-Pool GRPO, Reject-Sample SFT, and Offline-Augment GRPO baselines reach 71.75\%, 72.23\%, and 73.13\%, respectively.
The frozen-Synthesizer baseline improves further to 75.85\%, whereas \textsc{SynCo} reaches 78.72\%, outperforming the strongest external baseline by 3.76 percentage points, Frozen by 2.87 points, and Vanilla by 6.36 points.
\textsc{SynCo} achieves the best or tied-best result on six of the eight benchmarks and ranks second on the remaining two, indicating substantially more consistent improvements than methods that peak on individual datasets.
Relative to Frozen, the gains include 2.20 points on MATH-500, 2.05 points on GSM-Hard, and 1.67 points on SVAMP, while ASDiv is already near saturation.
On the advanced benchmarks, \textsc{SynCo} improves AIME24 and AIME25 by 6.66 points each and Minerva by 2.57 points over Frozen.
The overall results show that jointly adapting the synthesis policy yields a stronger and more broadly effective training distribution than existing synthetic-data approaches or a frozen Synthesizer.
The same pattern is reproduced with Qwen3-4B, where co-training yields consistent improvements over the frozen variant, as reported in Appendix~\ref{app:qwen4b}.

\subsection{RQ2: How Does the Synthesized Curriculum Evolve During Co-Training?}
\label{sec:rq2}

\begin{figure*}[t]
    \centering
    \includegraphics[width=\textwidth]{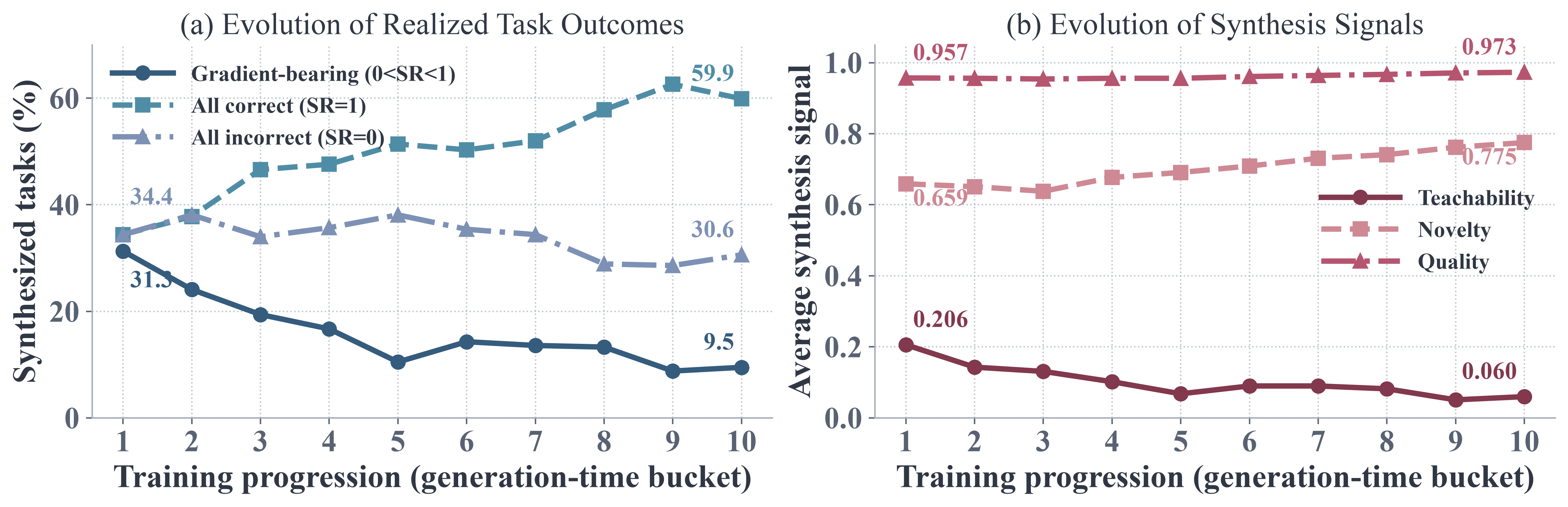}
    \caption{
Evolution of the synthesized curriculum during co-training.
Tasks are ordered by generation time and divided into ten equally sized buckets.
\textbf{(a)} Realized Reasoner outcomes, where gradient-bearing tasks have non-identical rollout rewards (equivalently, $0<SR<1$ under our reward setting).
\textbf{(b)} Evolution of synthesis signals, showing that task quality and novelty remain high while realized teachability declines.
Generation order approximates training progression because exact iteration indices were not recorded.
}
    \label{fig:rq2_synthesizer_evolution}
\end{figure*}

Figure~\ref{fig:rq2_synthesizer_evolution} reveals a progressive shift in the utility of the synthesized curriculum.
From the first to the final bucket, the proportion of gradient-bearing tasks decreases from 31.3\% to 9.5\%, while the all-correct proportion increases from 34.4\% to 59.9\%, indicating that an increasing fraction of generated tasks becomes uniformly solvable as the Reasoner improves.
Meanwhile, task quality remains consistently high and novelty increases from 0.659 to 0.775, whereas teachability declines sharply from 0.206 to 0.060.
The full-pool statistics in Appendix D.Table~\ref{tab:rq2_mechanism} further show that only 16.2\% of the 2,940 verified tasks provide non-degenerate group-relative feedback, and that teachability strongly tracks this gradient-bearing property ($r=0.934$), unlike novelty ($r=-0.041$).
These results distinguish task validity and diversity from realized optimization utility: well-formed and novel tasks may still provide no effective policy gradient for the current Reasoner.
Thus, although \textsc{SynCo} achieves consistent downstream gains, maintaining an informative curriculum becomes increasingly difficult as the learner evolves, motivating synthesis objectives grounded directly in realized rollout outcomes.
Additional analyses are in Appendix.D.

\subsection{RQ3: Where Do the Gains from Co-Training Arise?}
\label{sec:rq3}

To distinguish capability acquisition from preservation of existing knowledge, we partition each benchmark according to the initial Qwen3-8B predictions. As shown in Table~\ref{tab:rq3_gain_attribution}, \textsc{SynCo} and \textsc{SynCo} (Frozen Synth.) remain nearly identical on the Retain subset, whereas a much larger gap emerges on examples that the initial model fails to solve. In the pooled results, the Learn subset improves from 27.11\% to 32.23\%, while the Retain subset changes only from 98.52\% to 98.95\%. Correspondingly, 45 of the 56 additional solved examples arise from the Learn subset, showing that the advantage of co-training is dominated by capability acquisition rather than improved preservation of already mastered behavior. This pattern is consistent across all four benchmarks, indicating that adapting the synthesis policy does not merely reinforce the current policy but reallocates training experience toward regions where unresolved yet learnable problems remain. Together with the curriculum analysis in Section~\ref{sec:rq2}, these results suggest that learner-adaptive synthesis is particularly valuable near the evolving capability frontier, where informative training experience can be converted into new problem-solving competence. A complete benchmark-level decomposition is provided in Appendix~\ref{app:rq3} and Figure~\ref{fig:rq3_gain_attribution}.

\begin{table*}[t]
\centering
\scriptsize
\setlength{\tabcolsep}{3.8pt}
\renewcommand{\arraystretch}{1.08}
\caption{Instance-level attribution of the terminal performance difference between \textsc{SynCo} and the frozen-Synthesizer baseline. Examples are partitioned by the initial Qwen3-8B predictions: \emph{Retain} contains initially correct examples, whereas \emph{Learn} contains initially incorrect examples. Frozen and \textsc{SynCo} report the number and percentage of solved examples within each subset. $\Delta Q$ denotes the additional examples solved by \textsc{SynCo}, and \emph{Net} sums the contributions from both subsets.}
\label{tab:rq3_gain_attribution}
\begin{adjustbox}{max width=\textwidth}
\begin{tabular}{lrrrrrrrrr}
\toprule
\multirow{2}{*}{\textbf{Benchmark}} & \multicolumn{4}{c}{\textbf{Retain: Initially Correct}} & \multicolumn{4}{c}{\textbf{Learn: Initially Incorrect}} & \multirow{2}{*}{\textbf{Net $\Delta Q$}} \\
\cmidrule(lr){2-5}\cmidrule(lr){6-9}
& \textbf{$N$} & \textbf{Frozen} & \textbf{\textsc{SynCo}} & \textbf{$\Delta Q$} & \textbf{$N$} & \textbf{Frozen} & \textbf{\textsc{SynCo}} & \textbf{$\Delta Q$} & \\
\midrule
GSM-Hard & 683 & 661 (96.8\%) & \textbf{667 (97.7\%)} & $+6$ & 636 & 147 (23.1\%) & \textbf{168 (26.4\%)} & \textbf{$+21$} & \textbf{$+27$} \\
MATH-500 & 406 & 401 (98.8\%) & \textbf{403 (99.3\%)} & $+2$ & 94 & 27 (28.7\%) & \textbf{36 (38.3\%)} & \textbf{$+9$} & \textbf{$+11$} \\
GSM8K & 1,219 & 1,210 (99.3\%) & \textbf{1,212 (99.4\%)} & $+2$ & 100 & 40 (40.0\%) & \textbf{51 (51.0\%)} & \textbf{$+11$} & \textbf{$+13$} \\
SVAMP & 252 & 250 (99.2\%) & \textbf{251 (99.6\%)} & $+1$ & 48 & 24 (50.0\%) & \textbf{28 (58.3\%)} & \textbf{$+4$} & \textbf{$+5$} \\
\midrule
\textbf{Pooled} & \textbf{2,560} & \textbf{2,522 (98.52\%)} & \textbf{2,533 (98.95\%)} & \textbf{$+11$} & \textbf{878} & \textbf{238 (27.11\%)} & \textbf{283 (32.23\%)} & \textbf{$+45$} & \textbf{$+56$} \\
\bottomrule
\end{tabular}
\end{adjustbox}
\end{table*}

\subsection{RQ4: Which Synthesis Signals Align with Gradient-Bearing Tasks?}
\label{sec:rq4}

\begin{figure*}[t]
    \centering
    \includegraphics[width=\columnwidth]{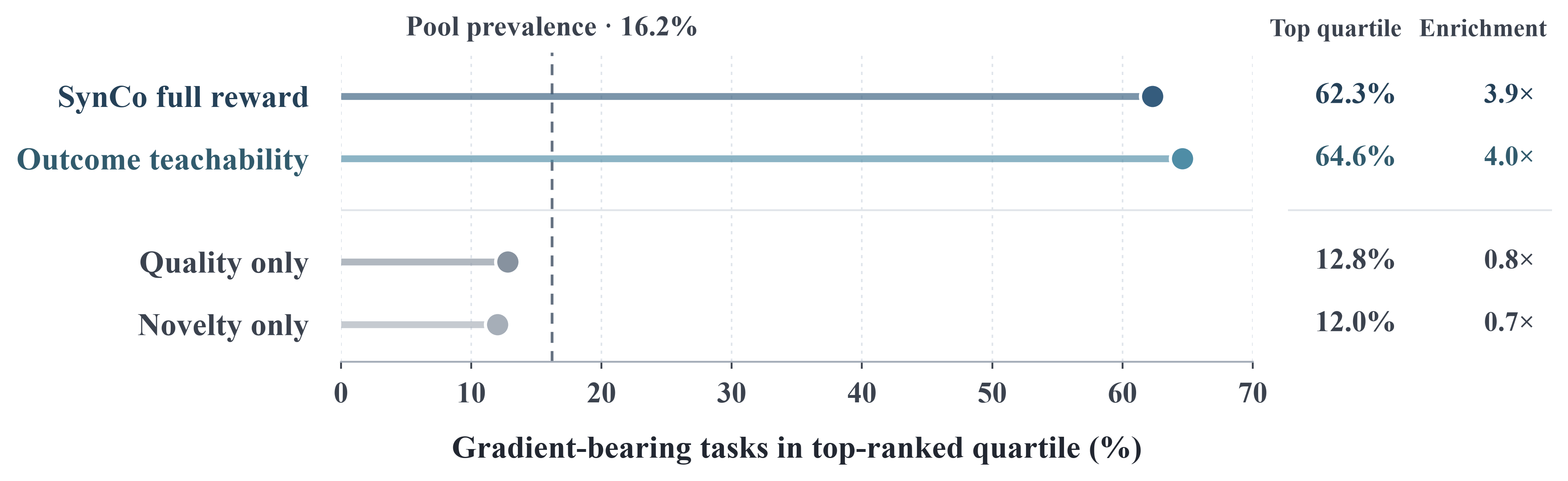}
    \caption{
    Counterfactual enrichment analysis of synthesis signals.
    Tasks are reranked offline using different signals, and the proportion of gradient-bearing tasks in the top-ranked quartile is reported.
    Outcome-grounded teachability strongly enriches for gradient-bearing tasks, whereas static task-level signals provide substantially weaker separation.
    }
    \label{fig:rq4_reward_enrichment}
\end{figure*}

Figure~\ref{fig:rq4_reward_enrichment} shows that outcome-grounded teachability is the synthesis signal most strongly aligned with optimization utility.
Relative to the 16.2\% gradient-bearing prevalence in the full task pool, ranking by the \textsc{SynCo} gated reward increases this proportion to 62.3\%, while teachability alone reaches 64.6\%.
In contrast, quality and novelty alone yield only 12.8\% and 12.0\%, respectively, both below the pool prevalence.
The signal-level statistics in Table~\ref{tab:rq4_reward_diagnostics} further show that teachability strongly correlates with the gradient-bearing indicator ($r=0.934$), whereas static task-level signals exhibit weak alignment.
These results separate task feasibility from learning utility: quality and reliability help ensure trustworthy tasks, but the Reasoner's realized rollout outcomes determine whether a task provides informative group-relative feedback.
The gated reward therefore preserves the strong outcome signal while constraining optimization to valid and reliable tasks, supporting the central design of \textsc{SynCo}.
Because this analysis reranks an already collected task pool, it should be interpreted as a diagnostic analysis rather than a causal training ablation; additional results are provided in Appendix~\ref{app:rq4}.


\subsection{RQ5: Can the Synthesizer Anticipate Gradient Utility Before Rollout?}
\label{sec:rq5}

\begin{figure*}[t]
    \centering
    \includegraphics[width=\columnwidth]{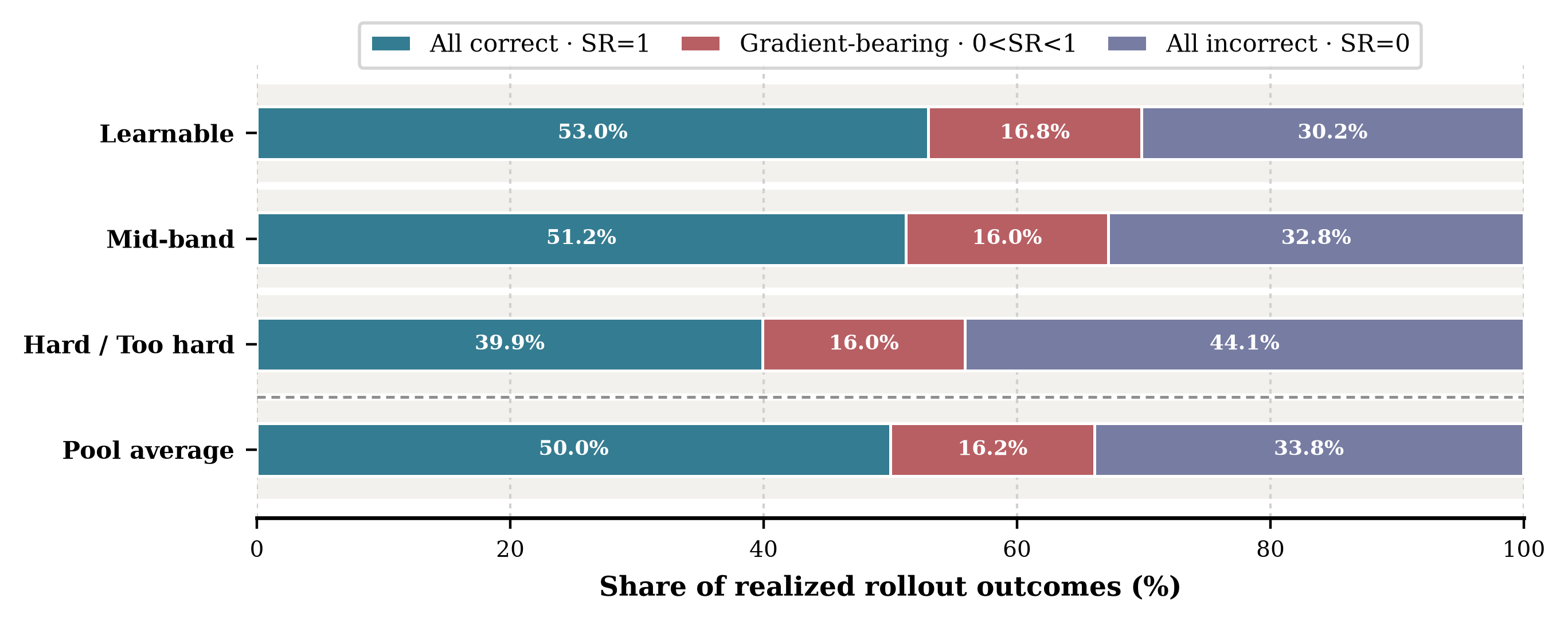}
    \caption{
    Realized Reasoner outcomes under the Synthesizer's ex-ante difficulty estimates.
    Harder labels shift tasks from uniformly correct toward uniformly incorrect outcomes, indicating coarse difficulty awareness.
    However, the gradient-bearing proportion remains nearly unchanged across difficulty categories, showing that nominal difficulty does not predict realized optimization utility.
    }
    \label{fig:rq5_difficulty_calibration}
\end{figure*}

Figure~\ref{fig:rq5_difficulty_calibration} shows that the Synthesizer's ex-ante estimates capture coarse task difficulty but fail to identify tasks with useful realized learning signal.
As the label shifts from \emph{learnable} to \emph{hard/too hard}, the all-correct rate decreases from 53.0\% to 39.9\%, while the all-incorrect rate increases from 30.2\% to 44.1\%.
However, this shift occurs almost entirely between the two zero-gradient extremes: the gradient-bearing rate remains nearly constant at 16.8\%, 16.0\%, and 16.0\% across the three major difficulty groups, closely matching the 16.2\% pool prevalence.
Table~\ref{tab:rq5_synthesizer_calibration} further shows that treating the \emph{learnable} label as a predictor yields only $r=0.009$, 16.8\% precision, and 22.3\% recall.
In contrast, the ex-post teachability signal studied in Section~\ref{sec:rq4} reaches $r=0.934$ after observing the Reasoner's rollouts.
Thus, the Synthesizer can broadly anticipate whether tasks will be easier or harder, but cannot reliably identify the intermediate region that produces non-degenerate learning signals.
This gap shows that nominal difficulty alone cannot substitute for learner interaction, supporting the outcome-grounded feedback design of \textsc{SynCo}.
Additional calibration results are provided in Appendix~\ref{app:rq5}.

\subsection{RQ6: How Efficient Is Online Synthesizer--Reasoner Co-Training?}
\label{sec:rq6}

\begin{table}[t]
\centering
\small
\setlength{\tabcolsep}{4.2pt}
\renewcommand{\arraystretch}{1.08}
\caption{
Training cost and gradient utilization for Frozen Synthesizer and \textsc{SynCo}.
Times are measured per training step, and productive rate denotes the fraction of gradient-bearing tasks.
}
\label{tab:rq6_efficiency_main}
\begin{tabular}{lcccc}
\toprule
\textbf{Variant} & \textbf{Update} & \textbf{Wall} & \textbf{Grad. Rate} & \textbf{Tasks/Batch} \\
\midrule
Frozen & 45.8\,s & 341.6\,s & 13.7\% & 4.38 \\
\textsc{SynCo} & 45.1\,s & 343.5\,s & \textbf{14.5\%} & \textbf{4.64} \\
\midrule
$\Delta$ & $-1.5\%$ & $+0.6\%$ & $+0.8$ pp & $+0.26$ \\
\bottomrule
\end{tabular}
\end{table}

Table~\ref{tab:rq6_efficiency_main} shows that jointly optimizing the Synthesizer introduces negligible marginal overhead relative to the matched frozen-Synthesizer pipeline.
The mean update time remains essentially unchanged (45.1\,s vs.\ 45.8\,s), while median end-to-end time differs by only 0.6\%.
Under this comparable cost, \textsc{SynCo} increases the productive-task rate from 13.7\% to 14.5\%, corresponding to 4.64 versus 4.38 productive tasks per batch.
This indicates that the benefit of co-training comes from reallocating the training distribution toward more informative interactions rather than simply increasing computation or rollout volume.
Because both variants share the same Reasoner optimization budget and per-task rollout configuration, the comparison isolates the effect of adapting the synthesis policy itself.
The higher productive-task rate further suggests that this adaptation improves how efficiently the fixed interaction budget is converted into useful learning signal.
Thus, the gains from co-training are not explained by additional per-step computation, but by a more effective online training distribution.
Complete timing and data-production statistics are provided in Appendix~\ref{app:rq6}.


\subsection{RQ7: Does SynCo Better Match the Learner's Capability Boundary?}
\label{sec:rq7}

\begin{figure*}[t]
    \centering
    \includegraphics[width=\textwidth]{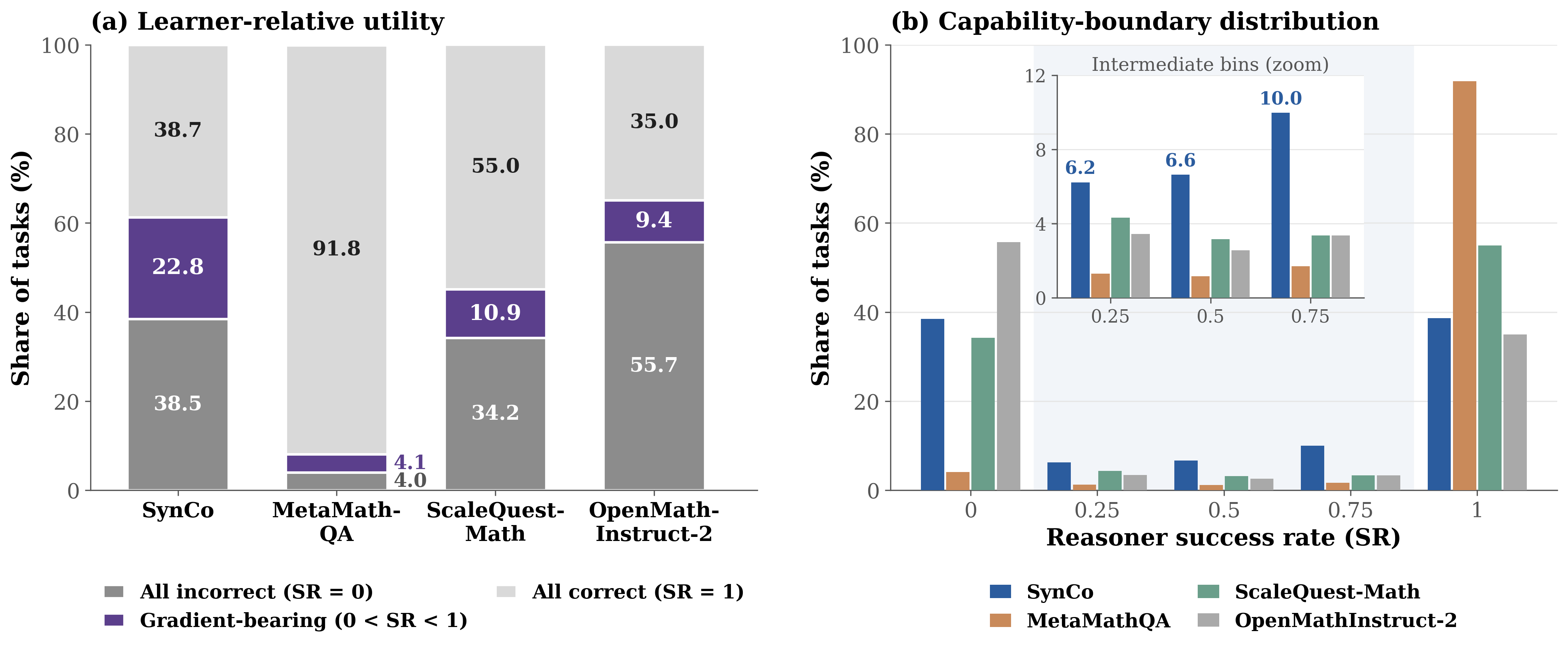}
    \caption{
    Learner-relative comparison of synthetic data from \textsc{SynCo}, MetaMathQA, ScaleQuest-Math, and OpenMathInstruct-2 under the same Vanilla Qwen3-8B Reasoner and rollout protocol.
    \textbf{(a)} Distribution of realized rollout outcomes, highlighting the proportion of gradient-bearing tasks.
    \textbf{(b)} Full Reasoner success-rate distribution, showing how each data source is positioned relative to the learner's capability boundary.
    }
    \label{fig:rq8_data_comparison}
\end{figure*}

\begin{figure}[t]
    \centering
    \includegraphics[width=0.80\columnwidth]{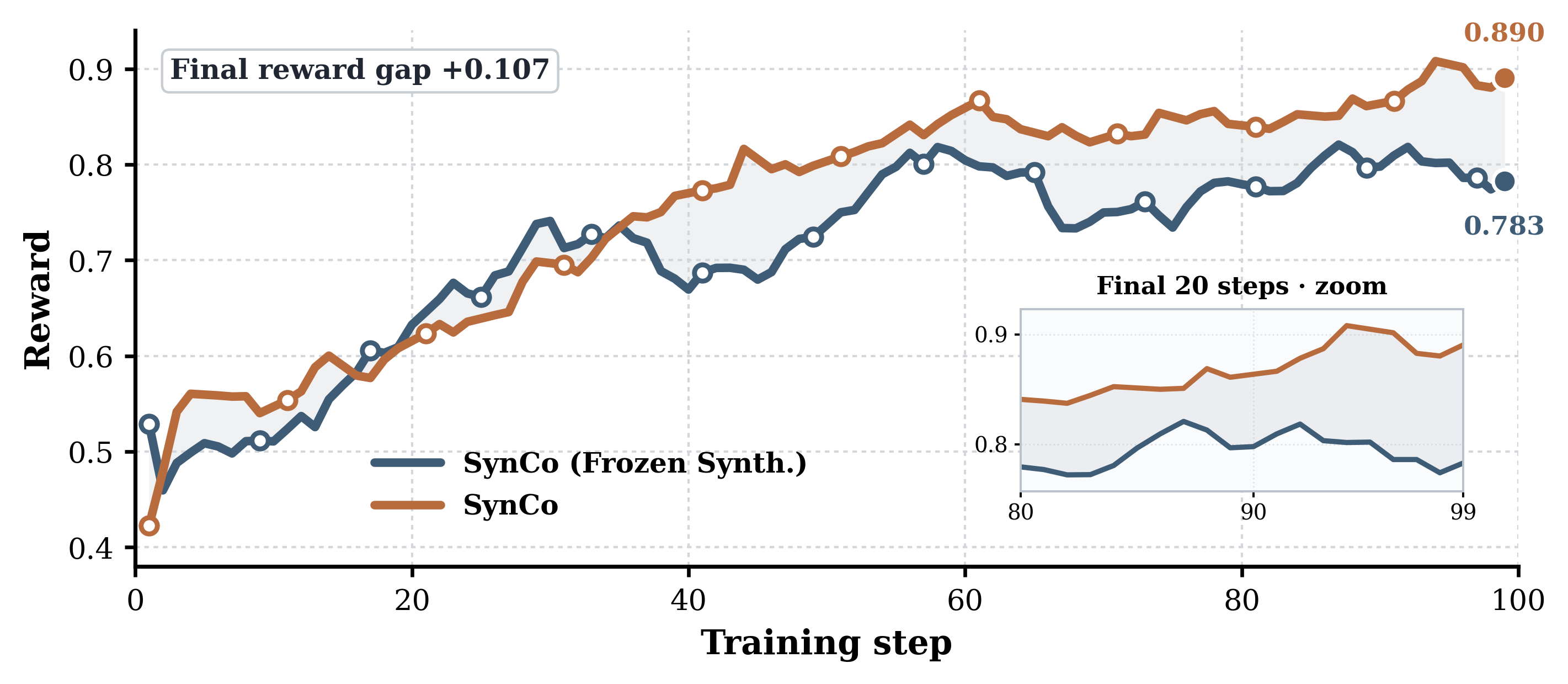}
    \caption{
    Reasoner reward trajectories under a frozen Synthesizer and joint Synthesizer--Reasoner co-training.
    Curves report 10-step moving averages.
    After comparable early progress, \textsc{SynCo} maintains a consistently higher reward trajectory.
    }
    \label{fig:rq8_reward_curve}
\end{figure}

Figure~\ref{fig:rq8_data_comparison} compares the learner-relative utility of \textsc{SynCo} data with three representative static synthetic-data sources under the same Vanilla Qwen3-8B Reasoner, rollout budget, and answer checker.
\textsc{SynCo} places 22.8\% of its tasks in the gradient-bearing region, compared with 4.1\% for MetaMathQA, 10.9\% for ScaleQuest-Math, and 9.4\% for OpenMathInstruct-2, corresponding to a $2.1$--$5.5\times$ increase in potentially informative group-relative interactions.
Panel (b) further shows that this advantage does not arise simply from generating harder problems: MetaMathQA is overwhelmingly concentrated at $\mathrm{SR}=1$, whereas OpenMathInstruct-2 places substantially more mass at $\mathrm{SR}=0$, and both therefore contain relatively few intermediate-outcome tasks.
In contrast, \textsc{SynCo} distributes considerably more probability mass across $\mathrm{SR}\in\{0.25,0.5,0.75\}$, indicating stronger concentration near the current learner's capability boundary.
These results provide a data-side explanation for the downstream gains in Section~\ref{sec:rq1}: the benefit of learner-conditioned synthesis is not merely producing difficult or diverse problems, but generating more training experiences that are appropriately matched to the current Reasoner and yield non-degenerate learning signal.

\subsection{RQ8: How Does Co-Training Affect the Reasoner's Optimization Dynamics?}
\label{sec:rq8}

Figure~\ref{fig:rq8_reward_curve} compares the Reasoner's optimization trajectories under the two training regimes.
Both variants improve rapidly at the beginning, indicating that online synthesized tasks already provide useful training signal with a frozen Synthesizer.
The trajectories begin to separate after approximately 35 steps, after which \textsc{SynCo} maintains a consistently higher moving-average reward.
By the end of training, \textsc{SynCo} reaches 0.890 compared with 0.783 for the frozen-Synthesizer baseline.
This sustained divergence shows that the benefit of co-training emerges progressively as the synthesis policy adapts to the evolving Reasoner, rather than only appearing at a particular evaluation checkpoint.
Together with the downstream gains in Section~\ref{sec:rq1}, these dynamics provide process-level evidence that joint Synthesizer--Reasoner optimization produces a more effective training trajectory.

\section{Conclusion}

In this paper, we introduced \textsc{SynCo}, a multi-agent framework that jointly optimizes task synthesis and reasoning through shared online interaction.
Across eight mathematical benchmarks, \textsc{SynCo} consistently improves over a frozen Synthesizer, with gains primarily arising from previously unsolved problems.
Our analyses further show that static task properties such as quality or novelty are insufficient to identify useful training experience, whereas outcome-grounded teachability is strongly aligned with gradient-bearing interactions throughout training.
Co-training achieves these gains with little measurable marginal overhead, establishing adaptive data synthesis as a learnable and scalable component of self-evolving LLM training.

\bibliographystyle{unsrtnat}
\bibliography{citation}

\clearpage
\appendix


\section*{Appendix}

\section{Related Work}
\label{app:related_work}

Large language models have increasingly been extended into autonomous agents capable of planning, tool use, memory, and multi-step interaction with external environments~\citep{shinn2023reflexion,guo2025deepseek,xu2026mem,zhang2026oases,ping2025verimoa,qin2026memory,yang2026ramem,yang2026auditing}.
Building on these capabilities, multi-agent systems coordinate multiple specialized agents through communication, collaboration, and iterative feedback, and have been explored across reasoning, coding, web interaction, and other complex decision-making tasks~\citep{zhang2024codeagent,he2024webvoyager,chang2025survey,li2026fortis,yang2025toward,yang2026adaptive,chen2026self,ping2026poet,yang2026tournament}.
These developments motivate learning frameworks in which agents not only solve tasks, but also actively construct and adapt the experience used for subsequent training.

\subsection{Synthetic Data for LLM Post-Training}

Synthetic data has become an important source of supervision for post-training large language models~\citep{gao2026self}.
Early instruction-synthesis methods, including Self-Instruct~\citep{wang2023self}, WizardLM~\citep{xu2024wizardlm}, and instruction backtranslation~\citep{li2023self}, bootstrap instruction--response pairs from seed demonstrations, teacher models, or unlabeled corpora.
Subsequent work improves the realism, diversity, and coverage of generated data.
MATRIX~\citep{tang2025synthesizing} constructs realistic post-training scenarios through multi-agent simulation, while FAC Synthesis~\citep{li2026less} explicitly targets missing features in the model's representation space.
Other approaches employ iterative refinement, filtering, and autonomous data engineering to better align synthetic corpora with downstream learning objectives~\citep{cao2025condor,tu2025resofilter}.
These studies establish data generation as an important dimension of LLM post-training, but typically construct an offline corpus that is subsequently consumed by a separately optimized learner.

Reasoning-oriented synthesis places additional emphasis on solution quality and answer verifiability.
STaR~\citep{zelikman2022star} bootstraps reasoning traces from successful model generations, while WizardMath~\citep{luo2025wizardmath} and MetaMath~\citep{yu2023metamath} expand mathematical supervision through instruction evolution and multi-perspective problem augmentation.
MAmmoTH~\citep{yue2024mammoth} combines natural-language and program-based rationales, whereas ToRA~\citep{gou2024tora} constructs tool-integrated reasoning trajectories.
DeepSeekMath~\citep{shao2024deepseekmath}, NuminaMath~\citep{li2024numinamath}, and related mathematical post-training pipelines further demonstrate the effectiveness of large-scale curated or synthesized reasoning corpora paired with outcome verification.
More recently, SwS~\citep{liang2026sws} identifies persistent learner weaknesses and synthesizes new problems targeting the corresponding concepts.
Together with reinforcement learning from verifiable rewards, these methods enable scalable reasoning supervision without requiring process-level human preference annotations.

Synthetic experience has also become central to training tool-using and interactive agents.
Explorer~\citep{pahuja2025explorer} synthesizes diverse multimodal web trajectories through large-scale environment exploration, while OS-Genesis~\citep{sun2025genesis} derives executable GUI tasks from successful interaction trajectories.
AutoPlay~\citep{ramrakhya2025scaling} explores application states before generating environment-grounded tasks; Trajectory2Task~\citep{wang2026trajectory2task} transforms valid tool trajectories into verifiable user-facing tasks; and WebAggregator~\citep{wang2026webaggregator} constructs search tasks requiring compositional evidence aggregation.
EigenData~\citep{chen2026eigendata} synthesizes multi-turn tool-use dialogues together with executable checkers, whereas DreamGym~\citep{chen2025scaling} uses a learned experience model to scale synthetic tasks and interactions.
These pipelines improve feasibility, environmental grounding, diversity, and verifiability, but their generators are commonly fixed, prompt-optimized, or updated separately from the policy trained on the resulting experience.

In group-relative reinforcement learning, however, the value of a task is not determined by validity, novelty, or nominal difficulty alone.
Even a well-formed task provides no discriminative policy signal when all sampled solutions are correct or all are incorrect, and its learning utility changes as the Reasoner evolves.
\textsc{SynCo} therefore formulates data synthesis as a trainable policy conditioned on the current learner state.
By optimizing the Synthesizer through verified downstream interaction, the training distribution can evolve according to the realized teachability of generated tasks rather than only their static data properties.

\subsection{Self-Evolving LLMs and Agent--Data Co-Training}

Self-evolving LLMs seek to reduce dependence on externally constructed supervision by generating and learning from their own experience~\citep{cheng2025evolving}.
R-Zero~\citep{huang2025r} initializes independently parameterized Challenger and Solver policies that co-evolve through task generation and solution feedback.
Socratic-Zero~\citep{wang2025socratic} extends this paradigm with Teacher, Solver, and Generator roles, using diagnosed learner weaknesses to construct an increasingly challenging mathematical curriculum.
Agent0~\citep{xia2025agent0} further incorporates tool-integrated reasoning into curriculum--executor co-evolution, allowing the generated tasks to advance with the executor's capabilities.
These methods demonstrate that task proposal and task solving can form an autonomous feedback loop rather than relying on a fixed, human-authored problem distribution.

Related frameworks extend self-evolution beyond standard mathematical reasoning.
Dr. Zero~\citep{yue2026dr} co-evolves a question proposer and a multi-turn search agent while reducing rollout cost through hop-grouped relative optimization.
SCOPE~\citep{kwan2026scope} trains a Challenger and retrieval-based Solver on open-ended, document-grounded tasks, using a frozen self-judge to construct task-specific evaluation rubrics.
CoEvolve~\citep{yang2026coevolve} extracts forgetting and uncertainty signals from agent trajectories to guide subsequent tool-use task synthesis.
Agent-World~\citep{dong2026agent}, SEAD~\citep{dai2026sead}, AgentEvolver~\citep{zhai2025agentevolver}, EvolveR~\citep{wu2025evolver}, and EvoFSM~\citep{zhang2026evofsm} similarly evolve environments, curricula, interaction experiences, memories, or agent workflows across tool use, service dialogue, search, and general agentic settings.
Collectively, this literature establishes agent--data feedback as a promising basis for autonomous and continual improvement.

The closest prior approaches adapt a proposer, teacher, or curriculum policy according to an evolving solver, commonly emphasizing frontier difficulty, weakness coverage, uncertainty, diversity, or verifier-based task quality.
Such adaptation is often implemented through round-wise curriculum reconstruction, stage-wise policy optimization, or a fixed generation module augmented with learner feedback.
\textsc{SynCo} instead adopts a more tightly coupled formulation: the same online interaction batch produces rollout-level correctness rewards for the Reasoner and task-level, interaction-grounded rewards for the Synthesizer, after which both independently parameterized policies are updated within a unified multi-agent training process.
This setting also differs from multi-agent post-co-training methods such as MAPoRL~\citep{park2025maporl}, which jointly optimize multiple agents for collaborative response generation rather than training one policy to construct another policy's future learning experience.
Beyond final performance, we explicitly analyze whether synthesized tasks are gradient-bearing, thereby distinguishing coarse difficulty and static data quality from the realized optimization utility observed under the current Reasoner.

\section{Method}
\label{app:method}
\subsection{Problem Formulation: Agentic Data Synthesis Co-Training}
\label{app:problem_formulation}

We consider a self-evolving learning system composed of two trainable agents.
The \emph{Synthesizer} is parameterized by $\theta_S$ and defines a data-synthesis policy
$\pi_S(\cdot;\theta_S)$, while the \emph{Reasoner} is parameterized by $\theta_R$ and defines a task-solving policy
$\pi_R(\cdot;\theta_R)$.
The two agents have fully separated parameters and are optimized for different but mutually dependent objectives.
The Synthesizer controls which training experiences are presented to the Reasoner, whereas the Reasoner's behavior on those experiences determines the task-level feedback received by the Synthesizer.

At training iteration $t$, the system maintains an evolving synthesis context $c_t$ that summarizes the current learning state.
Conceptually, $c_t$ captures the Reasoner's recent task performance, capability weaknesses, the coverage of the synthetic task distribution, and feedback from previously rejected or redundant tasks.
Given this context, the Synthesizer samples a structured task
\begin{equation}
    x_t \sim \pi_S(\cdot \mid c_t; \theta_S^t),
    \label{eq:synth_policy}
\end{equation}
where $x_t$ contains a problem specification together with the information required to verify its validity and answer.
Accordingly, the training distribution at iteration $t$ is not externally fixed, but is induced by the current synthesis policy and the current learner state:
\begin{equation}
    \mathcal{D}_t
    =
    \pi_S(\cdot \mid c_t; \theta_S^t).
    \label{eq:dynamic_distribution}
\end{equation}

For each valid synthesized task $x_t$, the Reasoner produces a group of $K$ independent solution rollouts,
\begin{equation}
    y_t^{(k)}
    \sim
    \pi_R(\cdot \mid x_t; \theta_R^t),
    \qquad
    k = 1,\ldots,K.
    \label{eq:reasoner_rollout}
\end{equation}
Let $a_t^\star$ denote the verified reference answer associated with $x_t$.
An outcome verifier evaluates each rollout and returns
\begin{equation}
    z_t^{(k)}
    =
    \operatorname{Check}
    \left(
        y_t^{(k)}, a_t^\star
    \right),
    \label{eq:outcome_verification}
\end{equation}
where $z_t^{(k)}$ indicates whether the rollout successfully solves the task.
The resulting interaction record is
\begin{equation}
    \mathcal{I}_t
    =
    \left(
        c_t,
        x_t,
        \left\{y_t^{(k)},z_t^{(k)}\right\}_{k=1}^{K}
    \right).
    \label{eq:interaction_record}
\end{equation}

The same interaction $\mathcal{I}_t$ induces agent-specific rewards for both policies.
For the Reasoner, each rollout receives a task-solving reward
\begin{equation}
    r_{R,t}^{(k)}
    =
    R_R
    \left(
        x_t,
        y_t^{(k)},
        z_t^{(k)}
    \right),
    \label{eq:reasoner_reward_general}
\end{equation}
which encourages correct and properly formatted solutions.
For the Synthesizer, the task receives a synthesis reward
\begin{equation}
    r_{S,t}
    =
    R_S
    \left(
        c_t,
        x_t,
        \left\{y_t^{(k)},z_t^{(k)}\right\}_{k=1}^{K}
    \right).
    \label{eq:synth_reward_general}
\end{equation}
Unlike a static task-quality score, $r_{S,t}$ depends on how the \emph{current} Reasoner actually behaves on the synthesized task.
It therefore evaluates not only whether the task is valid and reliable, but also whether it provides useful training experience for the present Reasoner.

The two policies optimize their respective expected returns,
\begin{align}
    J_R(\theta_R)
    &=
    \mathbb{E}_{x_t \sim \pi_S,\,
    y_t^{(k)} \sim \pi_R}
    \left[
        \frac{1}{K}
        \sum_{k=1}^{K}
        r_{R,t}^{(k)}
    \right],
    \label{eq:reasoner_objective}
    \\
    J_S(\theta_S)
    &=
    \mathbb{E}_{x_t \sim \pi_S,\,
    \{y_t^{(k)}\}_{k=1}^{K} \sim \pi_R}
    \left[
        r_{S,t}
    \right].
    \label{eq:synth_objective}
\end{align}
Although the agents receive distinct rewards and do not share parameters, their optimization problems are intrinsically coupled.
The distribution underlying $J_R$ is determined by the Synthesizer through Eq.~\eqref{eq:dynamic_distribution}, while the reward underlying $J_S$ is determined partly by the Reasoner's rollout distribution.
We jointly update the two policies from the same online interaction process:
\begin{align}
    \theta_R^{t+1}
    &=
    \mathcal{U}_R
    \left(
        \theta_R^t,
        \mathcal{I}_t,
        \{r_{R,t}^{(k)}\}_{k=1}^{K}
    \right),
    \label{eq:reasoner_update}
    \\
    \theta_S^{t+1}
    &=
    \mathcal{U}_S
    \left(
        \theta_S^t,
        \mathcal{I}_t,
        r_{S,t}
    \right),
    \label{eq:synth_update}
\end{align}
where $\mathcal{U}_R$ and $\mathcal{U}_S$ denote the corresponding group-relative policy optimization updates.

This formulation differs fundamentally from training a task model on an independently synthesized dataset.
Here, the Synthesizer directly controls the Reasoner's online learning distribution, and the Reasoner's realized behavior directly supervises the Synthesizer.
Consequently, data synthesis and task learning form a coupled multi-agent policy optimization problem rather than two sequential training stages.
The following sections instantiate this formulation by describing how the evolving Reasoner state conditions task synthesis, how shared task interactions jointly optimize both agents, and how their updates continuously reconstruct the subsequent training distribution.

\subsection{Reward and Verification Details}
\label{app:reward_details}

Each synthesized task is verified before entering Reasoner training.
The verifier checks schema completeness, problem well-posedness, answer consistency, output validity, and redundancy with recently generated tasks.
We additionally apply an independent consensus check to filter tasks whose reference answers are insufficiently supported by independent solutions.
Throughout training and evaluation, Reasoner answers are judged using the same checker, which supports normalized exact, numerical, symbolic, and set- or tuple-level equivalence.

For the Reasoner, each rollout receives a correctness-based reward: 1.0 for a correct answer in the prescribed format, 0.9 for a correct answer recovered without that format, and 0 otherwise.
For the Synthesizer, the main experiments use an outcome-grounded gated reward:
\begin{equation}
    r_{S,t}^{\mathrm{imm}}
    =
    G_q\!\left(r_t^{\mathrm{quality}}\right)
    G_r\!\left(r_t^{\mathrm{reliability}}\right)
    T(\mu_t),
\end{equation}
where the quality and reliability gates follow
\begin{equation}
    G(v)
    =
    \operatorname{clip}
    \left(
        \frac{v-0.5}{0.5},
        0,1
    \right).
\end{equation}
The task-quality score is a deterministic composite of verifier signals measuring schema completeness, problem validity, answer consistency, output format, and redundancy.
The reliability score specifically reflects the internal consistency between the Synthesizer-provided reference solution and final answer.
Both are computed using rule-based checks rather than an LLM judge and primarily serve as feasibility signals; exact verifier coefficients and thresholds are included in the supplementary implementation.

Learner-dependent utility is determined by realized teachability.
Given the Reasoner's empirical success rate
$\mu_t=\frac{1}{K}\sum_{k=1}^{K}z_t^{(k)}$,
we define
\begin{equation}
    T(\mu_t)
    =
    \begin{cases}
        2\mu_t, & \mu_t \leq 0.5,\\
        2(1-\mu_t), & \mu_t > 0.5.
    \end{cases}
\end{equation}
Teachability is therefore maximized when the Reasoner succeeds on half of its rollouts and decreases toward zero as the rollout group becomes uniformly correct or uniformly incorrect.
The gated objective consequently separates task feasibility from learner-relative utility: quality and reliability constrain optimization toward trustworthy tasks, while the current Reasoner's realized behavior determines whether those tasks provide informative group-relative feedback.

Tasks rejected by the verifier receive an immediate Synthesizer reward of $-0.5$ and are excluded from Reasoner training.
The training framework additionally applies a constant 0.1 bonus to non-empty model outputs.
Novelty and diagnostic-alignment signals are logged for analysis but do not directly contribute to the gated Synthesizer reward used in the main experiments.

\section{Experimental Details}
\label{app:experiments}

\subsection{Benchmarks}
\label{app:benchmarks}

We evaluate mathematical reasoning across eight benchmarks spanning elementary word problems, competition mathematics, and higher-level quantitative reasoning. GSM8K~\citep{cobbe2021training} contains grade-school math word problems requiring multi-step arithmetic reasoning, while SVAMP~\citep{patel2021nlp} stresses robustness to variations in quantities and problem structure. ASDiv~\citep{miao2020diverse} provides a more diverse collection of arithmetic word problems with heterogeneous linguistic and reasoning patterns. GSM-Hard~\citep{gao2023pal} increases the numerical difficulty of GSM8K-style problems while largely preserving their underlying reasoning structure. MATH-500~\citep{hendrycks2021measuring,lightman2024let} covers competition-level algebra, geometry, number theory, counting, probability, and related topics, requiring more involved symbolic and multi-step reasoning. AIME24 and AIME25 each contain problems from the corresponding American Invitational Mathematics Examination and provide a challenging evaluation of competition-style reasoning with exact answers. Minerva~\citep{lewkowycz2022solving} contains higher-level quantitative problems spanning advanced mathematics and scientific reasoning. Together, these benchmarks cover a broad difficulty range, from relatively saturated elementary reasoning to substantially more challenging problems with greater learning headroom. We use the standard test splits and report answer accuracy under the same answer-extraction and verification pipeline for all methods.

\subsection{Baselines}
We compare \textsc{SynCo} with three representative offline alternatives, an iterative self-evolution baseline, and a matched frozen-Synthesizer baseline.
\emph{Static-Pool GRPO} constructs a synthetic task pool before training and optimizes only the Reasoner using GRPO~\citep{shao2024deepseekmath}.
It isolates whether reinforcement learning over fixed synthetic data can match an adaptively evolving training distribution.
\emph{Offline-Augmented GRPO} constructs an offline task pool through mathematical question rewriting and variation inspired by MetaMath~\citep{yu2023metamath}.
The Reasoner is subsequently trained with GRPO while the augmented task distribution remains fixed.
\emph{Rejection-Sampling SFT} samples Reasoner trajectories, retains those with verified correct answers, and fine-tunes the model on the resulting successful solutions.
This baseline follows the generate--filter--fine-tune self-training paradigm of ReST$^{\mathrm{EM}}$~\citep{singh2023beyond}, without online policy optimization or adaptive synthesis.
\emph{R-Zero}~\citep{huang2025r} co-evolves a Challenger and a Solver by iteratively generating tasks near the Solver's capability boundary and optimizing the two roles through reinforcement learning.
Unlike \textsc{SynCo}'s same-batch joint optimization, R-Zero follows an alternating self-evolution process in which task generation and solver improvement are updated across successive stages.
Finally, \emph{\textsc{SynCo} (Frozen Synth.)} follows the same online generation and Reasoner-training pipeline as \textsc{SynCo}, but keeps the Synthesizer parameters fixed.
This matched comparison isolates the benefit of jointly optimizing the synthesis policy.

Further, we include two complementary groups of external baselines to provide a broader comparison with existing synthetic-data approaches.

\textbf{Released-checkpoint baselines.}
We first evaluate four publicly released models that have undergone synthetic-data-driven agentic post-training.
\textbf{AgenticQwen}~\citep{lyu2026agenticqwen} trains compact agentic models through multi-round reinforcement learning with dual data flywheels, where reasoning and agentic tasks are progressively synthesized and increased in difficulty.
\textbf{EnvScaler}~\citep{song2026envscaler} programmatically synthesizes executable tool-interactive environments together with scenarios and rule-based verification functions, and uses the resulting interaction data for SFT and RL of Qwen3 models.
\textbf{AgentSkiller}~\citep{sun2026agentskiller} automatically constructs executable cross-domain environments and synthesizes multi-turn interaction trajectories through structured domain modeling, semantic cross-domain composition, and execution-based validation.
\textbf{Klear-AgentForge}~\citep{wang2025klear} provides an open post-training pipeline that combines large-scale synthetic-data SFT with subsequent multi-turn reinforcement learning for general-purpose agentic capabilities.
For these baselines, we directly use the authors' publicly released checkpoints without additional training or adaptation, and evaluate them using exactly the same mathematical evaluation protocol as SynCo.

\textbf{Released synthetic-data baselines.}
We further compare against three representative publicly released mathematical synthesis corpora while controlling the downstream learner and optimization budget.
\textbf{MetaMathQA}~\citep{yu2024metamath} expands mathematical supervision through question rephrasing and backward-reasoning-based question generation, producing a large collection of augmented mathematical problems.
\textbf{ScaleQuest-Math}~\citep{ding2025unleashing} synthesizes mathematical questions from scratch using lightweight question generators optimized through Question Fine-Tuning and Question Preference Optimization, yielding a large-scale corpus of synthetic problem--solution pairs.
\textbf{OpenMathInstruct-2}~\citep{toshniwal2025openmathinstruct} constructs massive open mathematical supervision through problem and solution synthesis; for our comparison, we use its newly augmented synthetic problems rather than unchanged original benchmark questions.
For each corpus, we sample exactly 2,940 unique synthetic problems, matching the number of verified training tasks used by SynCo, and train the same Qwen3 learner using the same GRPO implementation, rollout configuration, optimizer settings, and total optimization budget as our controlled static-data baseline.

\subsection{Implementation Details}
\label{app:implementation_details}

We initialize the Synthesizer and the Reasoner as two independently parameterized copies of Qwen3-8B, and additionally repeat the main comparison with Qwen3-4B.
Both agents are optimized using GRPO through the UnityMAS-O~\citep{chen2026unitymas} training framework, with vLLM serving as the rollout backend.
Each training iteration contains 32 synthesized tasks, and the Reasoner samples $K=4$ independent trajectories for every admitted task.
We use an actor learning rate of $1\times10^{-6}$, a 10-step warm-up, weight decay of 0.1, and no KL penalty in the reward.
During training, the Synthesizer and Reasoner use temperatures of 1.0 and 0.9, respectively, with $p=0.95$ nucleus sampling.
The Qwen3-8B experiments are distributed across two 8-GPU worker groups, one for each agent.
The Qwen3-4B experiments follow the same training protocol and hyperparameters except for backbone-specific inference configurations. Each synthesized task is verified for validity and answer reliability before entering Reasoner training, with an additional consensus check used to filter unreliable reference answers.
For evaluation, we use temperature 0.6 and top-$p$ 0.95, and apply the same answer checker to all methods, supporting exact, numerical, symbolic, and set- or tuple-level equivalence.
The frozen-Synthesizer baseline uses the same backbone, Reasoner optimization, rollout budget, verifier, and evaluation protocol, while keeping the Synthesizer parameters fixed.

\lstdefinestyle{promptstyle}{
  basicstyle=\ttfamily\scriptsize,
  breaklines=true,
  columns=fullflexible,
  keepspaces=true,
  showstringspaces=false,
  frame=none
}

\newtcolorbox{agentpromptbox}[2][]{
  enhanced,
  breakable,
  colback=gray!3,
  colframe=gray!60,
  coltitle=black,
  fonttitle=\bfseries,
  fontupper=\small,
  title={#2},
  boxrule=0.5pt,
  arc=1mm,
  left=1.2mm,
  right=1.2mm,
  top=1mm,
  bottom=1mm,
  before skip=1.2mm,
  after skip=1.2mm,
  #1
}

\subsection{Agent Prompt Templates}
\label{app:agent_prompts}

We provide the prompt templates used by the Synthesizer and the Reasoner.
At each synthesis iteration, the placeholders in the Synthesizer user prompt are populated using the current synthetic-pool statistics, the Reasoner's recent capability diagnostics, and feedback from rejected tasks.
The Reasoner receives only the generated problem statement; all pedagogical metadata, reference solutions, and verification information are reserved for task validation and reward computation.

\begin{agentpromptbox}{Synthesizer Prompt}

\textbf{Role.}
The Synthesizer generates exactly one original, valid, and pedagogically useful mathematical reasoning problem at each synthesis iteration.
The generated task should target the current Reasoner's learning needs, remain independently verifiable, and provide a useful group-relative learning signal.

\textbf{System prompt.}
\begin{lstlisting}[style=promptstyle]
You are the Synthesizer in a two-agent mathematical self-evolution system.

Your task is to generate exactly one original, valid, and pedagogically useful mathematical reasoning problem for training a Reasoner model.

The generated task should:
1. be well-posed and self-contained;
2. have a unique and independently verifiable final answer;
3. require non-trivial, multi-step reasoning;
4. be reliable enough to serve as training data;
5. differ meaningfully from recently generated tasks;
6. target weaknesses or under-explored reasoning behaviors identified in the diagnostic reports; and
7. be likely to provide a useful learning signal for the current Reasoner.

Use the following broad capabilities as inspiration, but do not mechanically cycle through them or treat them as fixed categories:

- Multi-step arithmetic word reasoning:
  quantity tracking, rates, ratios, unit conversion, work and time,
  proportional reasoning, money or counting scenarios, and hidden
  intermediate quantities.

- Competition-style algebra:
  equations, inequalities, substitutions, systems, functional
  expressions, hidden constraints, and exact simplification.

- Number theory:
  divisibility, modular arithmetic, primes, remainders, parity,
  greatest common divisors, least common multiples, and
  Diophantine constraints.

- Counting, probability, and combinatorics:
  case analysis, arrangements, combinations, overcounting,
  conditional probability, and recurrence-like reasoning.

- Geometry without diagrams:
  angles, lengths, areas, similarity, coordinate geometry,
  circles, polygons, and algebraic geometric constraints.

- Symbolic and advanced reasoning:
  sequences, functions, polynomial identities, transformations,
  exact symbolic computation, and calculus-like manipulation
  when appropriate.

Important restrictions:
- Do not copy or paraphrase known benchmark questions.
- Do not mention benchmark names in the generated problem.
- Do not generate famous contest problems or standard textbook examples.
- Do not generate ambiguous or underspecified problems.
- Do not require diagrams, external knowledge, or subjective judgment.
- Do not include the solution or answer in the problem statement.
- Do not output multiple problems.
- Do not output anything except valid JSON matching the required schema.

Difficulty calibration:
The Reasoner is already capable of solving many standard mathematical
reasoning problems. Prefer tasks for which the Reasoner is likely to
succeed on some, but not all, independent attempts. Tasks that are
solved uniformly or failed uniformly provide little group-relative
learning signal. Avoid direct applications of a single standard
formula, single-step setups, or problems with an immediately obvious
solution procedure.

Prefer tasks involving at least one of the following:
- a non-obvious case distinction or boundary condition;
- a hidden constraint that must be checked after an initial solution;
- multiple interacting constraints;
- a subtle parity, indexing, or limiting-case consideration; or
- symbolic or combinatorial structure requiring careful manipulation.
\end{lstlisting}

\textbf{User prompt structure.}
\begin{lstlisting}[style=promptstyle]
Current synthetic pool report:
{pool_report}

Current Reasoner diagnostic report:
{reasoner_diagnostic_report}

Recent rejection summary:
{recent_rejection_summary}

Generate exactly one new mathematical reasoning task as a JSON object
with the following schema:

{output_schema}

Requirements:
- Output valid JSON only.
- Generate exactly one task.
- The problem must be original, well-posed, and self-contained.
- The problem must have a unique and verifiable final answer.
- The task should target the current Reasoner's learning needs rather
  than merely maximize difficulty.
- The problem statement must not reveal the solution or final answer.
\end{lstlisting}

\textbf{Output schema.}
\begin{lstlisting}[style=promptstyle]
{
  "teaching_intent": "...",
  "target_reasoning_style": "...",
  "expected_reasoner_weakness": "...",
  "difficulty_estimate": "...",
  "problem": "...",
  "solution": "...",
  "final_answer": "...",
  "verification_notes": "...",
  "novelty_notes": "..."
}
\end{lstlisting}

\end{agentpromptbox}

When no interaction history is available, the context fields are initialized with neutral descriptions indicating that no historical pool, Reasoner diagnostics, or rejection records are yet available.

\begin{agentpromptbox}{Reasoner Prompt}

\textbf{Role.}
The Reasoner solves the mathematical problem generated by the Synthesizer.
It receives only the problem statement and does not observe pedagogical metadata, reference solutions, verification notes, or other synthesis-time information.

\textbf{System prompt.}
\begin{lstlisting}[style=promptstyle]
You are the Reasoner in a mathematical reasoning training system.

Solve the given problem carefully.
Reason step by step.
Put the final answer inside \boxed{}.
Do not provide multiple final answers.
\end{lstlisting}

\textbf{User prompt structure.}
\begin{lstlisting}[style=promptstyle]
Problem:
{problem}

Solve the problem step by step and put the final answer inside \boxed{}.
\end{lstlisting}

\end{agentpromptbox}

\paragraph{Shared prompt conventions.}
Across the Synthesizer and Reasoner prompts, the system follows four conventions:
(1) the Synthesizer must generate exactly one original, well-posed, and independently verifiable task;
(2) the Synthesizer output must be valid JSON following an explicitly specified schema;
(3) task generation is conditioned on synthetic-pool statistics, Reasoner diagnostics, and rejection feedback when such interaction history is available; and
(4) the Reasoner receives only the generated problem statement, while pedagogical metadata, reference solutions, and verification information are reserved for task validation and reward computation.

\section{Extended Experimental Analysis}
\label{app:extended_experiments}

\subsection{Generalization Across Model Scales}
\label{app:qwen4b}

We further examine whether the benefits of synthesis co-training generalize across model scales by repeating the main comparison with Qwen3-4B.
As shown in Table~\ref{tab:rq1_4b_results}, the same progression observed with Qwen3-8B is reproduced: conditioning a frozen Synthesizer on the evolving Reasoner raises the average accuracy from 70.14\% to 73.72\%, while jointly optimizing the Synthesizer further improves it to 75.20\%.
Thus, co-training contributes an additional 1.48 percentage points over the matched frozen variant and 5.06 points over Vanilla.
\textsc{SynCo} improves over its frozen counterpart on seven of eight benchmarks and achieves the strongest overall average, outperforming EnvScaler and AgentSkiller by 4.43 and 11.83 points, respectively.
While EnvScaler attains stronger results on AIME24 and AIME25, \textsc{SynCo} performs more consistently across the broader benchmark suite, achieving the best result on five benchmarks and ranking second on the remaining three.
The improvement is therefore not driven by a single task family: gains appear across arithmetic reasoning, algebraic and competition-style benchmarks, and Minerva, with only a small regression on GSM-Hard.
These results reproduce the main Qwen3-8B finding at a smaller scale and show that the advantage of jointly adapting the synthesis policy is robust to the Reasoner backbone rather than being specific to a single model size.

\begin{table*}[t]
\centering
\scriptsize
\setlength{\tabcolsep}{3.3pt}
\renewcommand{\arraystretch}{1.06}
\caption{
Reasoning accuracy (\%) across eight mathematical reasoning benchmarks using Qwen3-4B.
Vanilla denotes the initial Qwen3-4B model, while \textsc{SynCo} (Frozen Synth.) retains learner-conditioned online synthesis but disables Synthesizer updates.
\emph{Avg.} is the unweighted average across all benchmarks.
Bold and underlined values indicate the best and second-best results, respectively.
}
\label{tab:rq1_4b_results}
\begin{tabular}{lccccccccc}
\toprule
\textbf{Method} & \multicolumn{5}{c}{\textbf{Primary Benchmarks}} & \multicolumn{3}{c}{\textbf{Advanced Benchmarks}} & \textbf{Avg.} \\
\cmidrule(lr){2-6}\cmidrule(lr){7-9}
& \textbf{GSM8K} & \textbf{SVAMP} & \textbf{ASDiv} & \textbf{GSM-Hard} & \textbf{MATH-500} & \textbf{AIME24} & \textbf{AIME25} & \textbf{Minerva} & \\
\midrule
Vanilla & 92.34 & 86.25 & 96.63 & 52.01 & 82.60 & 63.33 & 56.67 & 31.25 & 70.14 \\
\midrule
EnvScaler & 89.84 & 82.00 & 95.40 & 42.00 & 82.00 & \textbf{73.33} & \textbf{70.00} & 31.62 & 70.77 \\
AgentSkiller & 79.45 & 72.67 & 92.94 & 37.38 & 71.80 & 63.33 & 60.00 & 29.41 & 63.37 \\
\midrule
\textsc{SynCo} (Frozen Synth.) & \underline{93.10} & \underline{92.00} & \underline{97.78} & \textbf{61.33} & \underline{86.20} & 66.67 & 60.67 & \underline{31.99} & \underline{73.72} \\
\textsc{SynCo} & \textbf{93.71} & \textbf{93.00} & \textbf{98.28} & \underline{60.65} & \textbf{86.60} & \underline{70.00} & \underline{66.67} & \textbf{32.72} & \textbf{75.20} \\
\bottomrule
\end{tabular}
\end{table*}

\subsection{How Does the Synthesized Curriculum Evolve During Co-Training?}
\label{app:rq2}

Table~\ref{tab:rq2_mechanism} provides a detailed decomposition of the 2,940 verified tasks collected during co-training.
We define a task as \emph{gradient-bearing} when its Reasoner rollout group receives non-identical rewards and therefore produces non-zero group-relative advantages.
Empirically, this criterion is equivalent to $0<\mathrm{SR}<1$ for all verified task groups in our collected training data.
Only 475 tasks (16.2\%) satisfy this condition, compared with 994 (33.8\%) that are uniformly incorrect and 1,471 (50.0\%) that are uniformly correct.
Thus, the primary bottleneck is not producing valid tasks, but continually placing them in the region where the current Reasoner receives informative group-relative learning signal.

The temporal evolution in Figure~\ref{fig:rq2_synthesizer_evolution} shows that this region becomes increasingly difficult to maintain.
Gradient-bearing tasks decrease from 31.3\% in the earliest bucket to 9.5\% in the final bucket, while the all-correct proportion rises from 34.4\% to 59.9\%.
The shift is therefore driven primarily by tasks becoming uniformly solvable as the Reasoner improves, rather than by an increase in universally failed tasks.
Meanwhile, task quality remains high and novelty increases, showing that a curriculum can remain valid and diverse while progressively losing optimization value.
This divergence highlights the non-stationarity of synthetic-data utility: a task distribution that is useful for an earlier learner may gradually fall behind its evolving capability frontier.

Panel (b) further separates static synthesis properties from learner-dependent utility.
Novelty is nearly uncorrelated with the gradient-bearing indicator ($r=-0.041$), whereas outcome-grounded teachability is strongly aligned with it ($r=0.934$), as expected from its construction using realized rollout outcomes.
Teachability is also nearly uncorrelated with raw success rate ($r=-0.028$), indicating that it does not simply favor easier or harder tasks but instead emphasizes heterogeneous outcomes.
Panel (c) reaches the same conclusion from the Synthesizer's ex-ante perspective: the \emph{learnable}, \emph{mid-band}, and \emph{hard/too hard} groups exhibit gradient-bearing rates of 16.8\%, 16.0\%, and 16.0\%, respectively, all close to the 16.2\% pool prevalence.
Thus, intended difficulty meaningfully shifts whether tasks are solved or failed, but provides little information about whether they lie in the optimization-relevant region.
Together, these analyses show that validity, novelty, and nominal difficulty characterize different properties of synthetic data, but none substitutes for observing how the current learner actually interacts with it.

\begin{table*}[t]
\centering
\footnotesize
\setlength{\tabcolsep}{5.0pt}
\renewcommand{\arraystretch}{1.08}
\caption{
Mechanistic analysis of 2,940 verified tasks collected during co-training.
Empirically, gradient-bearing groups coincide exactly with $0<\mathrm{SR}<1$ across the collected training data.
Panel (a) summarizes realized rollout outcomes, Panel (b) relates synthesis signals to optimization utility, and Panel (c) compares the three major intended-difficulty groups with realized learnability; these groups cover 2,919 tasks.
}
\label{tab:rq2_mechanism}

\begin{tabularx}{\textwidth}{
    >{\raggedright\arraybackslash}X
    >{\raggedright\arraybackslash}X
    >{\centering\arraybackslash}p{0.18\textwidth}
    >{\centering\arraybackslash}p{0.18\textwidth}
}
\toprule

\multicolumn{4}{c}{\textbf{(a) Realized Rollout Outcomes}} \\
\midrule
\textbf{Outcome Regime}
& \textbf{Criterion}
& \textbf{Number of Tasks}
& \textbf{Percentage} \\
\midrule

All incorrect
& $\mathrm{SR}=0$
& 994
& 33.8\% \\

\textbf{Gradient-bearing}
& $\mathbf{0<\mathrm{SR}<1}$
& \textbf{475}
& \textbf{16.2\%} \\

All correct
& $\mathrm{SR}=1$
& 1,471
& 50.0\% \\

\midrule
\multicolumn{4}{c}{\textbf{(b) Synthesis Signals vs.\ Realized Utility}} \\
\midrule
\textbf{Synthesis Signal}
& \textbf{Correlation Target}
& \textbf{Correlation}
& \textbf{Observed Alignment} \\
\midrule

Teachability
& Gradient-bearing indicator
& \textbf{+0.934}
& Strong positive \\

Novelty
& Gradient-bearing indicator
& $-0.041$
& Negligible \\

Teachability
& Realized success rate
& $-0.028$
& Negligible \\

\midrule
\multicolumn{4}{c}{\textbf{(c) Intended Difficulty vs.\ Realized Learnability}} \\
\midrule
\textbf{Synthesizer Label}
& \textbf{Number of Tasks}
& \textbf{Gradient-Bearing Rate}
& \textbf{Zero-Gradient Rate} \\
\midrule

Learnable
& 630
& 16.8\%
& 83.2\% \\

Mid-band
& 1,926
& 16.0\%
& 84.0\% \\

Hard / Too hard
& 363
& 16.0\%
& 84.0\% \\

\bottomrule
\end{tabularx}
\end{table*}

\subsection{Attributing Gains to Capability Acquisition}
\label{app:rq3}

Table~\ref{tab:rq3_gain_attribution} decomposes the terminal advantage of \textsc{SynCo} over \textsc{SynCo} (Frozen Synth.) according to whether each evaluation example is initially solved by Qwen3-8B.
Both methods are evaluated at fixed terminal checkpoints, avoiding benchmark-specific checkpoint selection.
On the 2,560 initially correct examples, the two variants remain close: \textsc{SynCo} retains 2,533 examples (98.95\%), compared with 2,522 (98.52\%) for the frozen variant, yielding 11 additional retained solutions.
The difference is substantially larger on the 878 initially incorrect examples, where \textsc{SynCo} solves 283 (32.23\%) compared with 238 (27.11\%) for Frozen.
Thus, the Learn subset contributes 45 of the 56 additional solved examples, accounting for 80.4\% of the overall advantage.

\begin{figure}[t]
    \centering
    \includegraphics[width=\columnwidth]{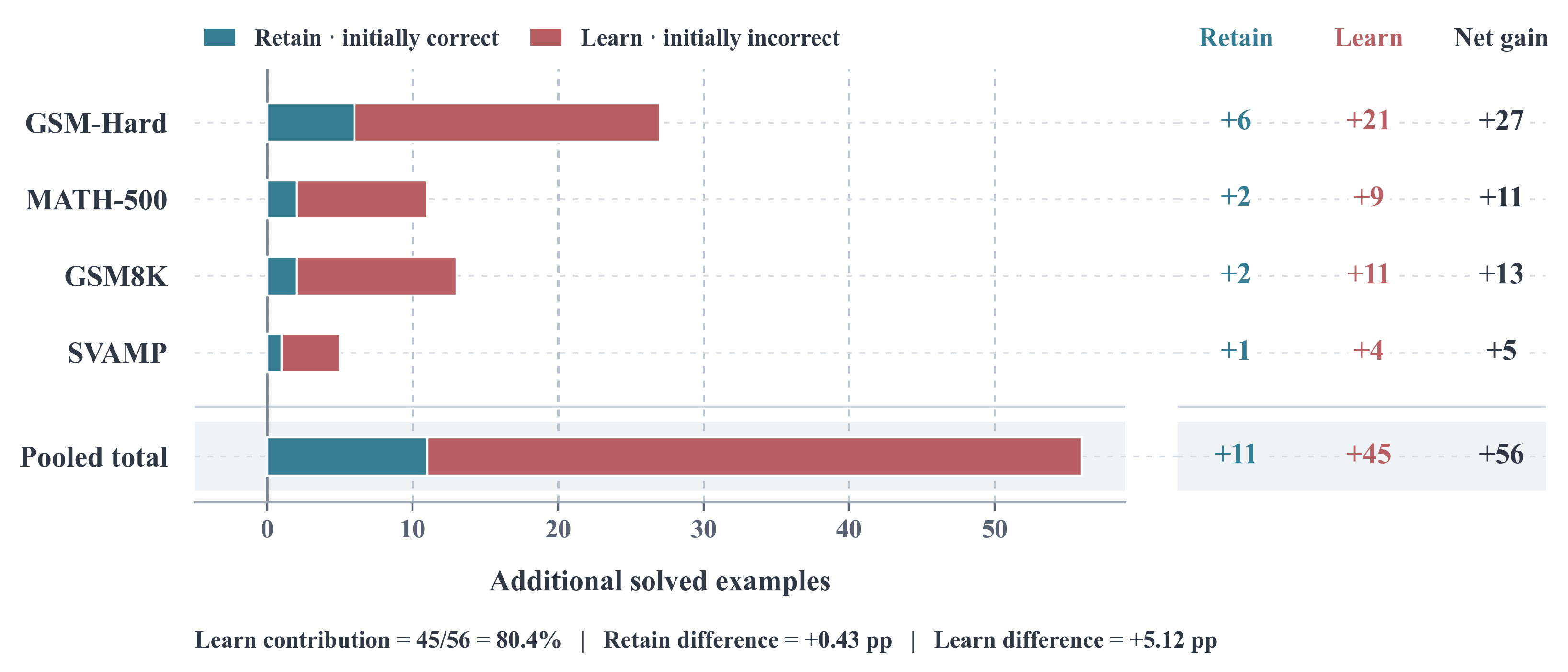}
    \caption{
Attribution of \textsc{SynCo}'s terminal advantage over \textsc{SynCo} (Frozen Synth.).
\emph{Retain} measures additional solved examples among those initially solved by Qwen3-8B, whereas
\emph{Learn} measures gains on initially incorrect examples.
Across all four benchmarks, most of the net improvement arises from the Learn subset, which contributes
45 of the 56 additional solved examples (80.4\%).
}
    \label{fig:rq3_gain_attribution}
\end{figure}

Figure~\ref{fig:rq3_gain_attribution} shows that this pattern holds consistently across all four benchmarks.
On GSM-Hard, the Learn subset contributes $+21$ of the $+27$ net gain, compared with $+6$ from Retain.
Similarly, Learn contributes $+9$ of $+11$ on MATH-500, $+11$ of $+13$ on GSM8K, and $+4$ of $+5$ on SVAMP.
In aggregate, co-training improves retention by only $+0.43$ percentage points, from 98.52\% to 98.95\%, while improving recovery on initially incorrect examples by $+5.12$ points, from 27.11\% to 32.23\%.
The terminal advantage is therefore driven primarily by increased success on examples outside the initial model's competence, rather than by preservation of already mastered behavior.

This decomposition complements the curriculum analysis in Section~\ref{sec:rq2}.
Although only a minority of synthesized tasks produce non-degenerate group-relative feedback, the resulting optimization translates predominantly into newly solved evaluation examples rather than additional retention of previously solved ones.
This pattern appears across all four benchmarks, including both relatively saturated datasets and benchmarks with greater unresolved headroom.
Together with the aggregate improvements in Section~\ref{sec:rq1}, these results support the role of adaptive synthesis in capability acquisition: jointly evolving the synthesis policy helps the Reasoner expand beyond its initial competence rather than merely consolidate existing performance.

\subsection{Synthesis Signals and Realized Gradient Utility}
\label{app:rq4}

Table~\ref{tab:rq4_reward_diagnostics} examines how different synthesis signals relate to realized optimization utility over the 2,940 verified tasks.
Quality and reliability are highly concentrated after verification, with means of 0.962 and 0.965, respectively.
This concentration is expected: both signals participate in the gated synthesis reward and are also closely related to the admissibility checks that remove malformed or unreliable tasks before Reasoner training.
Consequently, their weak correlations with the gradient-bearing indicator should not be interpreted as evidence that they are unnecessary; rather, once tasks pass verification, these signals provide limited discrimination over which remaining tasks will be useful for the current learner.
Novelty exhibits substantially greater variation but remains nearly uncorrelated with gradient production ($r=-0.041$), showing that a task can be distinct from previous generations without providing useful optimization signal.
Report alignment is similarly weakly associated with realized utility and is used as a diagnostic context signal rather than as part of the gated reward.

Outcome-grounded teachability behaves fundamentally differently.
Its correlation with the gradient-bearing indicator reaches $r=0.934$, while its correlation with raw Reasoner success rate is only $r=-0.028$.
Thus, teachability does not simply favor easier or harder tasks; it identifies tasks whose rollout groups contain heterogeneous outcomes and therefore provide non-degenerate group-relative feedback.
This distinction is central to the reward design of \textsc{SynCo}: quality and reliability constrain optimization toward trustworthy tasks, while realized Reasoner behavior determines which of those tasks are useful for learning.

The reranking analysis in Figure~\ref{fig:rq4_reward_enrichment} makes this separation explicit.
Selecting the top quartile according to the actual \textsc{SynCo} gated reward raises the gradient-bearing rate from the pool prevalence of 16.2\% to 62.3\%, a $3.9\times$ enrichment.
Outcome teachability alone reaches 64.6\% ($4.0\times$), whereas quality-only and novelty-only rankings yield only 12.8\% and 12.0\%, respectively.
The small difference between teachability alone and the full gated reward reflects their different roles: teachability identifies outcome-level learning utility, while the quality and reliability gates prevent optimization from favoring informative but invalid or unreliable tasks.
Hence, the objective is not designed to maximize gradient yield in isolation, but to identify useful learning experiences within a feasible synthesis space.

\begin{table*}[t]
\centering
\scriptsize
\setlength{\tabcolsep}{4.0pt}
\renewcommand{\arraystretch}{1.08}
\caption{
Diagnostic statistics of synthesis signals over 2,940 verified tasks.
Quality, reliability, and teachability constitute the gated reward used by \textsc{SynCo}, whereas novelty and report alignment are recorded as diagnostic signals.
Correlation is measured against the binary gradient-bearing indicator defined by the Reasoner's group-relative learning signal.
}
\label{tab:rq4_reward_diagnostics}

\begin{tabular*}{\textwidth}{
@{\extracolsep{\fill}}
l
c
c
c
p{3.0cm}
p{5.0cm}
@{}
}
\toprule
\textbf{Synthesis Signal}
& \textbf{Mean}
& \textbf{Std.}
& \shortstack{\textbf{Corr.\ with}\\\textbf{Gradient-Bearing}}
& \textbf{Role}
& \textbf{Interpretation within Verified Pool} \\
\midrule

Quality
& 0.962
& 0.030
& $+0.060$
& Reward gate
& Nearly saturated after verification, with limited discrimination among valid tasks. \\

Reliability
& 0.965
& 0.096
& $+0.105$
& Reward gate
& Ensures answer consistency and reliability, but weakly predicts realized gradient utility. \\

Report alignment
& 0.999
& 0.011
& $-0.021$
& Diagnostic
& Nearly constant in the collected pool and weakly associated with realized utility. \\

Novelty
& 0.703
& 0.175
& $-0.041$
& Diagnostic
& Varies substantially across tasks but does not identify gradient-bearing interactions. \\

\textbf{Teachability}
& \textbf{0.102}
& \textbf{0.249}
& \textbf{$+0.934$}
& \textbf{Outcome utility}
& \textbf{Strongly aligned with whether the task produces non-degenerate group-relative feedback.} \\

\bottomrule
\end{tabular*}
\end{table*}

\begin{table*}[t]
\centering
\scriptsize
\setlength{\tabcolsep}{3.2pt}
\renewcommand{\arraystretch}{1.05}
\caption{
Calibration of the Synthesizer's ex-ante difficulty estimates against realized Reasoner outcomes over 2,940 verified tasks.
Panel (a) compares self-rated difficulty with the outcomes of $K=4$ Reasoner rollouts, while Panel (b) contrasts the ex-ante \emph{learnable} label with outcome-grounded teachability as signals of realized gradient utility.
}
\label{tab:rq5_synthesizer_calibration}

\begin{tabular}{lccccc}
\toprule
\multicolumn{6}{c}{\textbf{(a) Ex-Ante Difficulty vs.\ Realized Rollout Outcomes}} \\
\midrule
\textbf{Difficulty Label} & \textbf{Tasks} & \textbf{Mean SR} &
\shortstack{\textbf{Gradient-}\\\textbf{Bearing}} &
\shortstack{\textbf{All}\\\textbf{Correct}} &
\shortstack{\textbf{All}\\\textbf{Incorrect}} \\
\midrule
Learnable & 630 & 0.625 & 16.8\% & 53.0\% & 30.2\% \\
Mid-band & 1,926 & 0.607 & 16.0\% & 51.1\% & 32.8\% \\
Hard / Too hard & 363 & 0.484 & 16.0\% & 39.9\% & 44.1\% \\
\midrule
\textbf{Pool overall} & \textbf{2,940} & \textbf{0.594} & \textbf{16.2\%} &
\textbf{50.0\%} & \textbf{33.8\%} \\
\bottomrule
\end{tabular}

\vspace{4pt}

\begin{tabularx}{\textwidth}{l >{\raggedright\arraybackslash}X c c c >{\raggedright\arraybackslash}X}
\toprule
\multicolumn{6}{c}{\textbf{(b) Ex-Ante vs.\ Outcome-Grounded Signals}} \\
\midrule
\textbf{Signal} & \textbf{Availability} &
\shortstack{\textbf{Corr.\ with}\\\textbf{Gradient-Bearing}} &
\textbf{Precision} & \textbf{Recall} & \textbf{Interpretation} \\
\midrule
Self-rated ``learnable'' & Ex ante, before rollout & $+0.009$ & 16.8\% & 22.3\% &
Captures generation intent but provides little enrichment for gradient-bearing tasks. \\
\textbf{Teachability} & Ex post, after rollout & \textbf{$+0.934$} & -- & -- &
\textbf{Outcome-grounded signal aligned with non-degenerate group-relative feedback.} \\
\bottomrule
\end{tabularx}
\end{table*}

\subsection{Why Ex-Ante Difficulty Does Not Predict Gradient Utility}
\label{app:rq5}

Table~\ref{tab:rq5_synthesizer_calibration} compares the Synthesizer's ex-ante difficulty estimates with the outcomes realized by the current Reasoner.
The estimates exhibit meaningful coarse calibration: tasks labeled \emph{learnable} have an average success rate of 0.625, whereas those labeled \emph{hard/too hard} average 0.484.
Consistently, the proportion of uniformly correct tasks decreases from 53.0\% to 39.9\%, while the uniformly incorrect proportion increases from 30.2\% to 44.1\%.
The Synthesizer therefore captures the broad direction of task difficulty and can shift generation from easier toward harder regions of the task space.

However, calibration of average difficulty does not imply calibration of optimization utility.
The gradient-bearing rates remain nearly unchanged at 16.8\%, 16.0\%, and 16.0\% for the \emph{learnable}, \emph{mid-band}, and \emph{hard/too hard} groups, respectively, all close to the 16.2\% pool prevalence.
Increasing nominal difficulty therefore primarily transfers probability mass from uniformly correct to uniformly incorrect outcomes rather than increasing the likelihood of mixed rollout groups.
This distinction is important under group-relative optimization: a task can become substantially harder while remaining equally uninformative if all sampled trajectories still receive the same outcome.
Ex-ante difficulty and realized gradient utility are therefore related but fundamentally different quantities.

Panel (b) quantifies this gap directly.
Treating the Synthesizer's \emph{learnable} label as a binary predictor of gradient-bearing tasks yields a correlation of only $0.009$, with 16.8\% precision and 22.3\% recall.
Its precision is essentially identical to the underlying pool prevalence, indicating that the label provides almost no enrichment for optimization-relevant tasks.
Moreover, most gradient-bearing tasks fall outside the \emph{learnable} category, showing that useful interactions cannot be isolated reliably through self-estimated difficulty alone.
The difficulty label is therefore better interpreted as a generation intent than as a prediction of downstream learning value.

By contrast, outcome-grounded teachability is computed after observing the Reasoner's rollouts and is strongly aligned with the gradient-bearing indicator ($r=0.934$) by construction.
Its role is consequently different from ex-ante difficulty: the former identifies realized learning utility, whereas the latter only guides where in the task space synthesis should occur before interaction.
Together with Section~\ref{sec:rq4}, this analysis explains why \textsc{SynCo} closes the loop between generation and learner feedback.
Static task attributes and self-estimated difficulty can steer exploration, but determining whether a task actually provides non-degenerate learning signal requires observing the current Reasoner's behavior.

\begin{table*}[t]
\centering
\scriptsize
\setlength{\tabcolsep}{4.5pt}
\renewcommand{\arraystretch}{1.08}
\caption{
Detailed efficiency analysis of online Synthesizer--Reasoner co-training.
Panel (a) compares per-step training cost, Panel (b) reports productive-task yield over the matched logging window, and Panel (c) summarizes task generation and verification during the 100-step \textsc{SynCo} run.
}
\label{tab:rq6_efficiency_full}

\begin{tabular}{lcccc}
\toprule
\multicolumn{5}{c}{\textbf{(a) Marginal Training Cost}} \\
\midrule
\textbf{Training Variant}
& \textbf{Steps}
& \shortstack{\textbf{Mean Update}\\\textbf{Time}}
& \shortstack{\textbf{Median Update}\\\textbf{Time}}
& \shortstack{\textbf{Median Wall}\\\textbf{Time}} \\
\midrule
\textsc{SynCo} (Frozen Synth.) & 100 & 45.8\,s & 45.0\,s & 341.6\,s \\
\textsc{SynCo} & 100 & 45.1\,s & 44.6\,s & 343.5\,s \\
\midrule
Relative difference & -- & $-1.5\%$ & $-0.9\%$ & $+0.6\%$ \\
\bottomrule
\end{tabular}

\vspace{6pt}

\begin{tabular}{lcccc}
\toprule
\multicolumn{5}{c}{\textbf{(b) Productive-Task Yield over the Matched Window}} \\
\midrule
\textbf{Training Variant}
& \textbf{Learnable}
& \textbf{Mid-band}
& \textbf{Productive}
& \shortstack{\textbf{Productive Tasks}\\\textbf{/ Batch}} \\
\midrule
\textsc{SynCo} (Frozen Synth.) & 3.4\% & 10.3\% & 13.7\% & 4.38 \\
\textsc{SynCo} & \textbf{3.9\%} & \textbf{10.6\%} & \textbf{14.5\%} & \textbf{4.64} \\
\midrule
Relative change & $1.14\times$ & $1.03\times$ & $1.06\times$ & $+0.26$ \\
\bottomrule
\end{tabular}

\vspace{6pt}

\begin{tabular}{lc}
\toprule
\multicolumn{2}{c}{\textbf{(c) Online Data Production}} \\
\midrule
\textbf{Quantity} & \textbf{Value} \\
\midrule
Generated task candidates & 3,200 \\
Verified tasks & 2,940 \\
Acceptance rate & 91.9\% \\
Human annotations & 0 \\
External LLM API calls & 0 \\
\bottomrule
\end{tabular}

\vspace{3pt}
\parbox{\textwidth}{
\footnotesize
Update time denotes the synchronized policy-update phase.
Wall-clock statistics are reported as secondary references because the two runs were executed independently.
Productive-task statistics are computed over steps 13--70, the interval with valid logging for both variants.
Task-generation, verification, and rollout computation are included in the reported runtime.
}
\end{table*}

\subsection{Training Efficiency, Gradient Utilization, and Data Production}
\label{app:rq6}

Table~\ref{tab:rq6_efficiency_full} provides a detailed efficiency analysis of \textsc{SynCo} and its frozen-Synthesizer counterpart.
As shown in Panel (a), jointly optimizing the Synthesizer introduces little additional per-step computation.
The mean synchronized update times are 45.1\,s for \textsc{SynCo} and 45.8\,s for Frozen, with corresponding median update times of 44.6\,s and 45.0\,s.
Median end-to-end step times are likewise comparable at 343.5\,s and 341.6\,s.
Because the runs were executed independently, these small wall-clock differences should be interpreted as comparable computational profiles rather than evidence that either variant is intrinsically faster.
The key result is that enabling Synthesizer optimization does not materially increase the cost of the otherwise matched online training pipeline.

Panel (b) examines how this comparable computation is converted into useful training signal over the matched logging window.
The productive-task rate increases from 13.7\% under Frozen to 14.5\% with \textsc{SynCo}, corresponding to 4.38 versus 4.64 productive tasks per batch of 32.
The increase is modest, but it is obtained without increasing the rollout budget or Reasoner optimization cost.
This suggests that adapting the synthesis policy improves the allocation of a fixed interaction budget: rather than generating more experience, \textsc{SynCo} slightly increases the frequency with which generated tasks fall into regions that provide non-degenerate group-relative feedback.
This complements the optimization dynamics in Section~\ref{sec:rq7}, where the same training process produces a progressively stronger Reasoner trajectory.

Panel (c) summarizes the resulting online data-production process.
Across 100 training steps, \textsc{SynCo} generates 3,200 task candidates, of which 2,940 pass verification, yielding a 91.9\% acceptance rate.
The resulting curriculum is generated and verified within the training loop without human annotation or external LLM API calls.
These numbers should not be interpreted as zero synthesis cost: generation, verification, and Reasoner rollouts are already included in the measured runtime.
Rather, they show that the evolving training distribution is constructed internally as part of the same optimization process, without requiring a separate data-generation stage.



\end{document}